\documentclass{article}

\usepackage[eandd, final]{neurips_2026}

\usepackage[utf8]{inputenc} 
\usepackage[T1]{fontenc}    
\usepackage{hyperref}       
\usepackage{url}            
\usepackage{graphicx}       
\usepackage{booktabs}       
\usepackage{amsfonts}
\usepackage{amsmath}       
\usepackage{nicefrac}       
\usepackage{microtype}      
\usepackage{xcolor}         
\usepackage{subcaption}

\title{Sanity Checks for Sparse Autoencoders: \\
 Do SAEs Beat Random Baselines?}

\author{%
  Anton Korznikov\thanks{Equal contribution. Correspondence: \texttt{korznikovantona@gmail.com}} \\
  AXXX \\
  HSE University \And
  Andrey Galichin\footnotemark[1] \\
  AXXX \\
  Applied AI Institute \And
  Alexey Dontsov \\
  AXXX \\
  HSE University \AND
  Oleg Rogov \\
  AXXX \\
  MTUCI Safe AI Lab \\
  Applied AI Institute \And
  Ivan Oseledets \\
  AXXX \And
  Elena Tutubalina \\
  AXXX \\
  HSE University
}

\begin{document}

\maketitle

\begin{abstract}

Sparse Autoencoders (SAEs) have emerged as a promising tool for interpreting neural networks by decomposing their activations into sparse sets of human-interpretable features. Despite much excitement, a growing number of negative results in downstream tasks casts doubt on whether SAEs recover meaningful features. To directly investigate this, we perform two complementary evaluations. On a synthetic setup with known ground‑truth features, we demonstrate that three out of four state‑of‑the‑art SAE architectures recover only 7–9\% of true features despite achieving around 71\% explained variance, showing that strong reconstruction alone is insufficient to guarantee meaningful feature recovery. To evaluate SAEs on real activations, we introduce three baselines that constrain SAE feature directions or their activation patterns to random values. Through extensive experiments across multiple SAE architectures, we show that our baselines match fully-trained SAEs in interpretability (0.87 vs 0.90), sparse probing (0.69 vs 0.72), and causal editing (0.73 vs 0.72). These results show that current evaluation metrics are insufficient to certify that SAEs have learned meaningful features, and we offer our baselines as a reusable protocol for future SAE evaluation.

\end{abstract}

\section{Introduction}

Large Language Models (LLMs) have achieved remarkable performance across a wide range of natural language processing tasks, with applications continuing to expand rapidly. As these models grow in capability and deployment, the interpretation of their internal mechanisms becomes increasingly important~\citep{bereska2024mechanistic, grace2025thousands}. Interpretability could potentially help to understand specific learning behaviors, such as safety mechanisms~\citep{arditi2024refusal} and reasoning processes~\citep{wang2022interpretability, galichin2025have}, identify misalignment risks~\citep{betley2025emergent, wang2025persona}. Sparse Autoencoders (SAEs) serve as a popular tool for this purpose, aiming to decompose dense model activations into sparse human-interpretable features~\citep{bricken2023monosemanticity, cunningham2023sparse}. Recent contributions have introduced various SAE architectures~\citep{rajamanoharan2024jumping, bussmann2024batchtopk, korznikov2025ortsae, bussmann2025learning}, with further success in scaling to frontier models~\citep{gao2024scaling,lieberum2024gemma}. However, a critical challenge remains: lacking ground truth for true features, \textit{it is difficult to determine whether SAEs recover meaningful representations.} Recent negative results on downstream applications~\citep{smithnegative, wu2025axbench} further underscore this challenge.

\begin{figure}[t]
  \centering
  \begin{subfigure}[b]{0.63\linewidth}
    \centering
    \includegraphics[width=\linewidth]{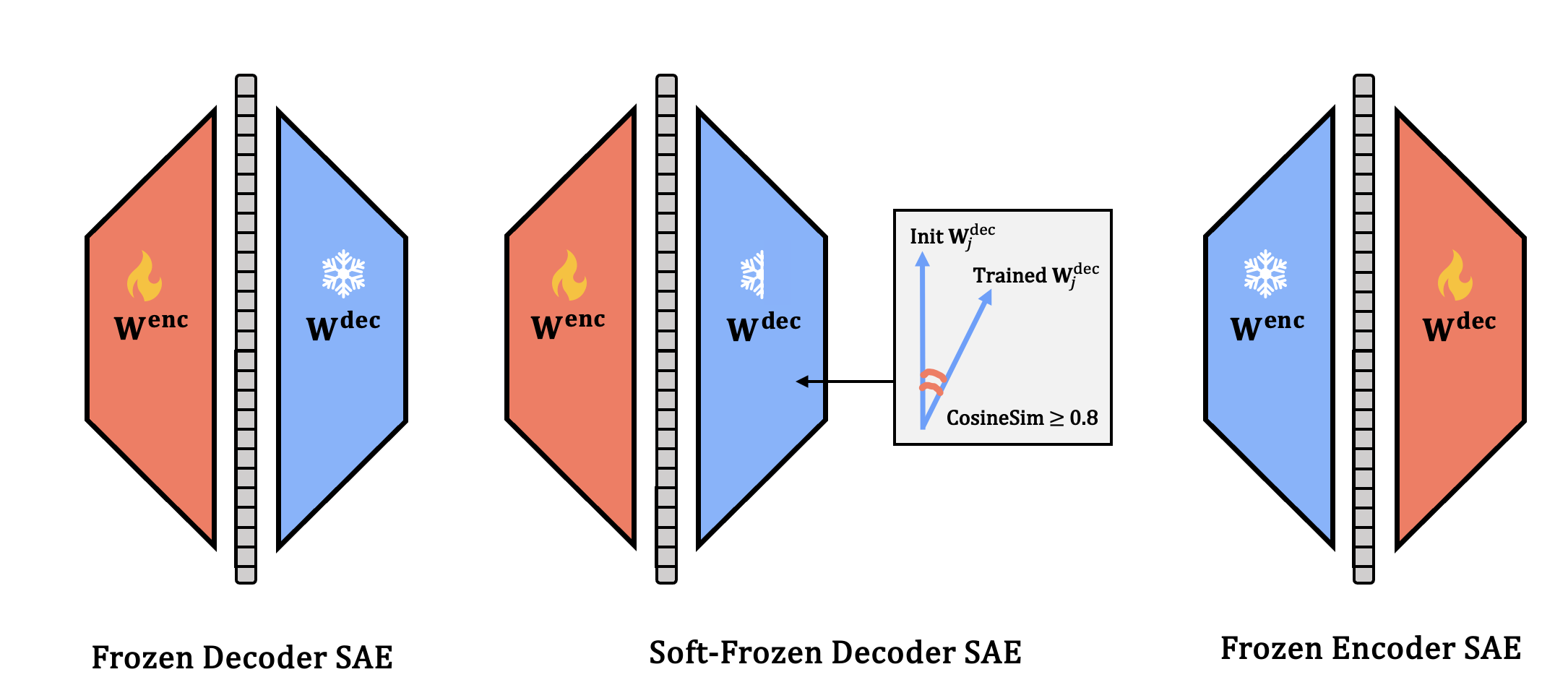}
  \end{subfigure}
  \hfill
  \begin{subfigure}[b]{0.35\linewidth}
    \centering
    \includegraphics[width=\linewidth]{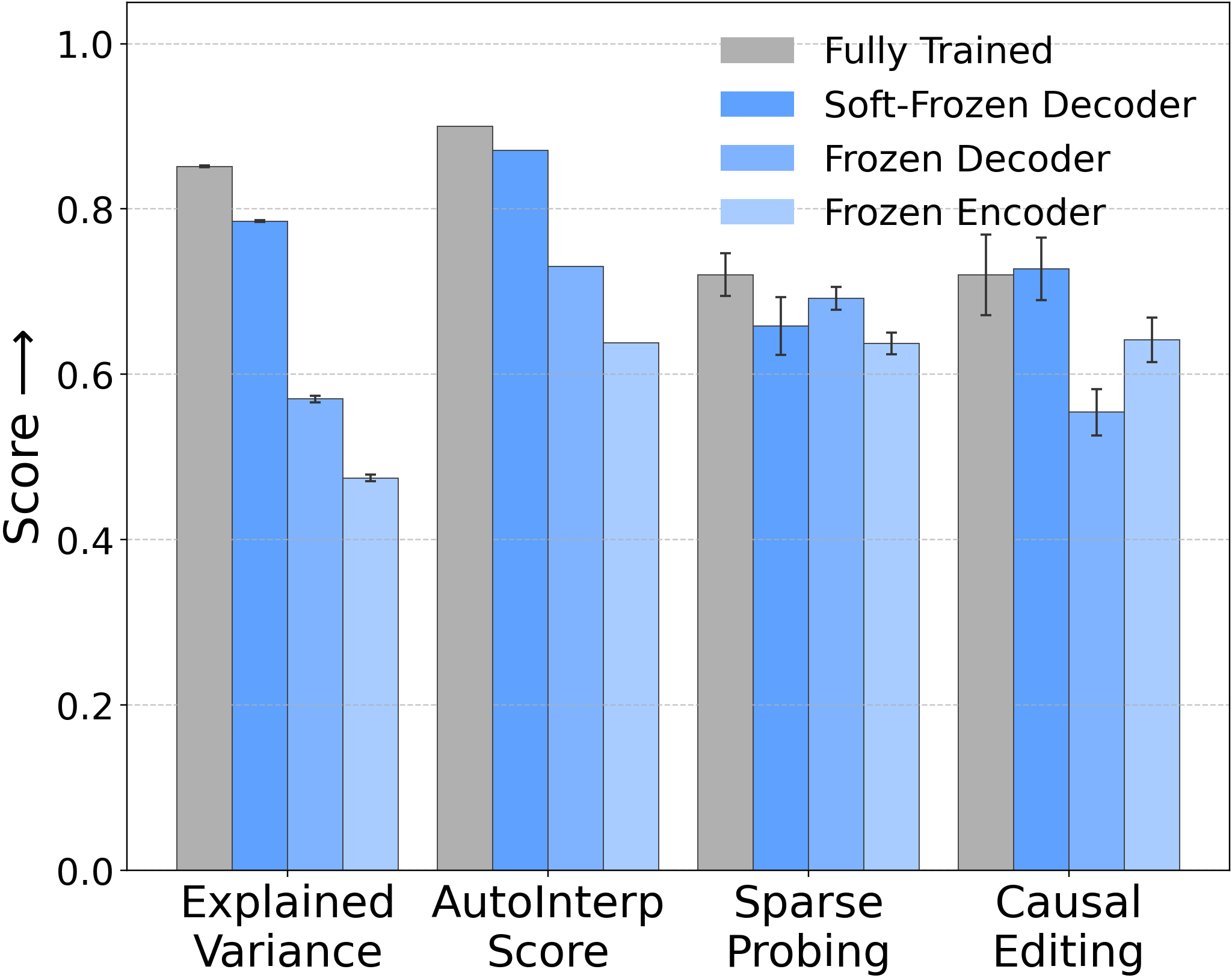}
  \end{subfigure}
  \caption{\textbf{Frozen SAE baselines and their performance.} (Left) Conceptual diagrams: Frozen Decoder SAE (decoder weights fixed at random initialization), Soft-Frozen Decoder SAE (decoder weights initialized randomly and constrained to maintain CosineSim $\geq$ 0.8 with their initial values throughout training), and Frozen Encoder SAE (encoder weights fixed at random initialization). (Right) For BatchTopK SAE (L0=160), these baselines remain competitive with fully trained SAE across four key evaluation metrics, challenging the assumption that strong performance indicates meaningful feature learning. Right-panel error bars are mean$\pm$std over 3 seeds (Appendix~\ref{sec:ablation}; AutoInterp is single-run).}
\label{fig:main_figure}
\end{figure}

To address this ambiguity, we test whether current metrics can certify meaningful feature learning. We first set up a synthetic experiment with known ground-truth features and test whether state-of-the-art SAE architectures can recover them. Despite achieving 71\% explained variance, 3 out of 4 SAE architectures recover only the highest-frequency features (7–9\% of total), failing to discover the decomposition they are intended to find. Second, to evaluate SAEs on real activations where ground truth is unknown, we compare them against three easy-to-implement baselines on downstream tasks. These baselines constrain SAE feature directions or activation patterns to random values (Figure ~\ref{fig:main_figure}): (1) \textbf{Frozen Decoder} fixes directions to random vectors; (2) \textbf{Frozen Encoder} fixes activation patterns to random; (3) \textbf{Soft-Frozen Decoder} restricts directions to remain near random initialization. The rationale is straightforward: the learned directions and activation patterns represent the learned decomposition. Consequently, if standard metrics reliably reflect meaningful feature learning, fully-trained SAEs should strongly outperform these baselines.

Across multiple SAE architectures, we find that our baselines match fully-trained SAEs on reconstruction fidelity~\citep{karvonen2025saebench} (0.79 vs 0.85), interpretability~\citep{paulo2024automatically} (0.87 vs 0.90), sparse probing~\citep{karvonen2025saebench} (0.69 vs 0.72), and causal editing~\citep{huang2024ravel} (0.73 vs 0.72). Together with our synthetic results, these findings suggest that current evaluations are insufficient, and we offer our baselines as sanity checks for validating whether SAEs learn meaningful feature decompositions.

Our work makes the following contributions:
\begin{itemize}

    \item We introduce a synthetic evaluation setup with known ground-truth features and demonstrate that SAE architectures often fail to recover them, revealing a disconnect between reconstruction fidelity and feature recovery.

    \item We propose three frozen baselines that constrain SAE components to random values, serving as a reusable evaluation protocol and minimal sanity check for future SAE work.


    \item We show that on real LLM activations, these baselines match fully-trained SAEs across all standard evaluation metrics - implying that current metrics are insufficient to certify meaningful feature decomposition.
\end{itemize}

\section{Background}

\subsection{Sparse Autoencoders for Interpretability}

\textbf{Model Architecture and Decomposition.} SAEs have emerged as a prominent tool for understanding neural network internals. It mainly addresses the challenge of polysemanticity, where individual neurons respond to multiple unrelated concepts \citep{bricken2023monosemanticity}. The core motivation behind SAEs is the \textit{superposition hypothesis} \citep{elhage2022toy}, which posits that neural networks represent more features than they have dimensions by encoding them as directions in activation space. Formally, for an activation vector \(\mathbf{x} \in \mathbb{R}^{n}\), this hypothesis states that \(\mathbf{x}\) can be expressed as a sparse linear combination of feature vectors:

\begin{equation}
\mathbf{x} = \sum_{j=1}^{m} a_j \cdot \mathbf{f}_j,
\end{equation}

where \(\{\mathbf{f}_j\} \subset \mathbb{R}^{n}\) are the underlying feature directions (with $m \gg n$) and \(a_j\) are sparse (mostly zero) nonnegative scalar coefficients. SAEs aim to learn this decomposition by approximating \(\mathbf{x}\) as:

\begin{equation}
\mathbf{x} \approx \hat{\mathbf{x}} = \sum_{j=1}^{m} z_j \cdot \mathbf{d}_{j} = \sum_{j=1}^{m} z_j \cdot \mathbf{W}^{\text{dec}}_{j} = \mathbf{W}^{\text{dec}} \mathbf{z},
\end{equation}

where \(\mathbf{d}_j \in \mathbb{R}^{n}\) are the learned decoder column vectors (the SAE's estimate of the true features \(\mathbf{f}_j\)), and \(z_j\) is a sparse activation produced by the encoder: \(\mathbf{z} = f(\mathbf{W}^{\text{enc}} \mathbf{x} + \mathbf{b}^{\text{enc}})\). Here \(f\) is a sparsity-inducing activation (e.g., ReLU) and \(\mathbf{b}^{\text{enc}}\) is a learned bias. The full reconstruction is \(\hat{\mathbf{x}} = \mathbf{W}^{\text{dec}} \mathbf{z} + \mathbf{b}^{\text{dec}}\). Typically, SAEs use an expansion factor \(k = m/n > 1\) (e.g. \(k \in \{16, 32, 64\}\)) to learn overcomplete dictionaries that can represent more features than the original dimensionality.

\textbf{Training Objective.} SAEs are trained to minimize a reconstruction loss while encouraging sparsity in the latent activations. The standard training objective combines mean squared error with an $L_1$ sparsity penalty:
\begin{equation}
\mathcal{L} = \mathbb{E}_{\mathbf{x}} \left[\|\mathbf{x} - \hat{\mathbf{x}}\|^2_2 + \lambda \|\mathbf{z}\|_1\right],
\end{equation}
where $\lambda$ controls the sparsity-reconstruction trade-off. Different SAE variants enforce sparsity through alternative mechanisms: some directly constrain the number of active features (measured by L0 norm, the count of non-zero elements in $\mathbf{z}$), while others use adaptive thresholds. A fundamental premise of SAEs is that optimizing both reconstruction and sparsity will yield learned feature directions ${\mathbf{d}_j}$ that align with the true underlying model features ${\mathbf{f}_j}$ and correspond to meaningful concepts.

Pioneering work by \citet{bricken2023monosemanticity} and \citet{cunningham2023sparse} demonstrated this approach on small transformers, discovering interpretable features such as DNA sequences and legal terminology. Subsequent efforts successfully scaled SAEs to frontier models, including Claude 3 Sonnet \citep{templeton2024scaling}, GPT-4 \citep{gao2024scaling}, and open-source models \citep{lieberum2024gemma}.

\subsection{SAE Architectural Variants}
Related work has proposed alternative SAE architectures to improve training stability and reconstruction quality. While vanilla SAEs use ReLU activation \citep{bricken2023monosemanticity, cunningham2023sparse}, \citet{rajamanoharan2024jumping} introduced JumpReLU, which adds learnable bias terms to activation thresholds, enabling dynamic sparsity adjustment during training. \citet{bussmann2024batchtopk} proposed BatchTopK, which enforces sparsity by selecting top-$k$ activations across batches rather than per sample, promoting feature reuse and potentially improving interpretability. Many more SAE variants have been proposed \citep{bussmann2025learning, korznikov2025ortsae, rajamanoharan2024improvingdictionarylearninggated}, but we focus mainly on the above-mentioned two variants due to their popularity and state-of-the-art performance on SAEBench \citep{karvonen2025saebench}.

\subsection{Critical Perspectives on SAEs}
Despite generating significant excitement, SAEs have accumulated a growing body of evidence documenting substantial limitations. While claimed benefits include decomposing activations into monosemantic features \citep{bricken2023monosemanticity} and successful scaling to frontier models \citep{templeton2024scaling, gao2024scaling}, recent work has identified these issues: SAEs fail to faithfully represent true model computations \citep{leask2025sparse}, show poor generalization across tasks \citep{heindrich2025sparse, kantamneni2025sparse}, learn features in corrupted ways like feature absorption \citep{chanin2025absorption} or feature hedging \citep{chanin2025hedging}, underperform on downstream applications \citep{wu2025axbench, smithnegative}, and exhibit high sensitivity to hyperparameters and initialization \citep{chanin2025sparse, paulo2025sparseautoencoderstraineddata, minegishi2025rethinking}. These failure modes are concerning because SAEs can yield high scores on standard evaluation metrics – reconstruction fidelity, interpretability, sparse probing – while failing to capture meaningful structure (a comprehensive overview is provided in Appendix~\ref{app:sae_overview}).

This tension between benefits and limitations motivates a fundamental question: are current evaluation metrics sufficient to certify that SAEs learn meaningful feature decompositions? Our work contributes to this literature by directly testing this assumption in both controlled synthetic settings and on real model activations.



\section{Case Study \#1: Validating SAEs on synthetic activations}
\label{sec:toy}

An implicit premise of SAE evaluation is that high reconstruction under a hard sparsity constraint signals recovery of the true features: with only a few active latents, any misalignment should severely degrade fidelity. We put this premise to the test in a controlled synthetic setting with known ground-truth features. Although prior work has validated SAEs on synthetic data \citep{chanin2025sparse, sharkey2022taking, elhage2022toy}, these studies typically used a small \textit{expansion factor} $k$ ($\approx$2), which simplifies the recovery problem and may not reflect the challenge faced in practice, where SAEs employ large expansion factors (e.g., 32 to 256) to learn highly overcomplete dictionaries. To bridge this gap, we use a realistic expansion factor of 32 and two data regimes, providing a more demanding test of the link between reconstruction quality and feature recovery.

\subsection{Experimental setup}
\label{sec:toy_setup}

\textbf{Synthetic Data Generation.} We base our toy model on the \textit{superposition hypothesis} \citep{elhage2022toy}, which posits that neural network activations can be represented as a sparse sum of interpretable feature vectors. To test whether SAEs can recover such features, we generate synthetic activations $\mathbf{x} \in \mathbb{R}^{n}$ (with $n = 100$) by first sampling an overcomplete dictionary of $3200$ ground‑truth feature vectors $\{\mathbf{f_i}\}$, each drawn uniformly from the unit sphere $S^{n-1}$. Each sample $\mathbf{x}$ is then a sparse combination of these features, expressed by:
\begin{equation}
\mathbf{x} = \sum_{i=1}^{3200} b_i \cdot c_i \cdot \mathbf{f}_i
\end{equation}
Here, \(c_i \sim \text{Log-Normal}(0, 0.25)\) governs the coefficient magnitude, while \(b_i \sim \text{Bernoulli}(p_i)\) determines whether a feature is active.  We examine two settings for the activation probability \(p_i\): the \textit{Constant Probability Model} with \(p_i = 0.00625\) across all features, and the \textit{Variable Probability Model} where \(p_i \sim \text{Log-Uniform}(10^{-5.5}, 10^{-1.2})\), reflecting the long-tailed activations found in practice \citep{lieberum2024gemma}. Both variants yield an expected sparsity of 20 active features.

\textbf{SAE variants.} 
We evaluated four state-of-the-art SAE architectures, BatchTopK \citep{bussmann2024batchtopk}, JumpReLU \citep{rajamanoharan2024jumping}, TopK \citep{gao2024scaling}, and Matryoshka \citep{bussmann2025learning} on this synthetic dataset. The SAE dictionary size was set to $3200$ with target L0 = 20, matching the ground truth.

\textbf{Evaluation.} 
Following \citet{gao2024scaling, rajamanoharan2024jumping, bussmann2024batchtopk}, we measured reconstruction fidelity using explained variance:
\begin{equation}
\text{Explained Variance} = 1 - \frac{\mathbb{E}\left[\|\mathbf{x} - \hat{\mathbf{x}}\|_2^2\right]}{\mathbb{E}\left[\|\mathbf{x} - {\mathbb{E}[\mathbf{x}}]\|_2^2\right]}
\label{eq:1}
\end{equation}
This metric ranges from 0 (reconstruction no better than the mean prediction) to 1 (perfect reconstruction).

Following \citet{sharkey2022taking}, to quantify feature recovery, we compute the cosine similarity between each ground-truth feature and its nearest SAE latent:
\begin{equation}
\text{Recovery}(\mathbf{f}_i) = \max_{j} \frac{\langle\mathbf{f}_i , \mathbf{W}^{\text{dec}}_j\rangle}{||\mathbf{f}_i||_2 \cdot||\mathbf{W}^{\text{dec}}_j||_2}
\end{equation}

High similarity scores indicate successful alignment between the learned and true features.

\subsection{Results}
\label{toy_model_res}

\begin{figure}[t]
  \centering
  \begin{subfigure}[b]{0.49\linewidth}
    \centering
  \includegraphics[width=1\columnwidth]{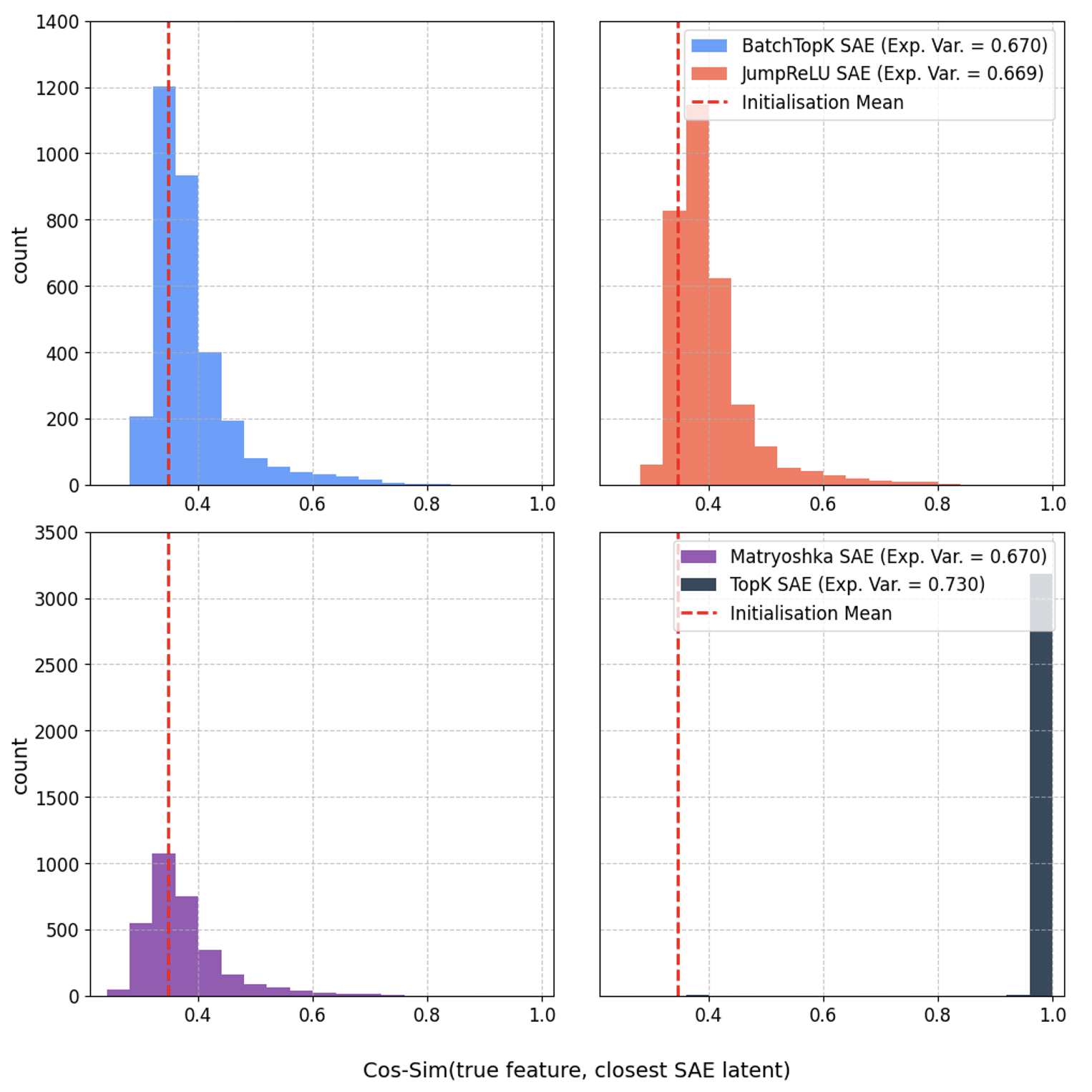}
\end{subfigure}
\begin{subfigure}[b]{0.49\linewidth}
  \centering
  \includegraphics[width=1\columnwidth]{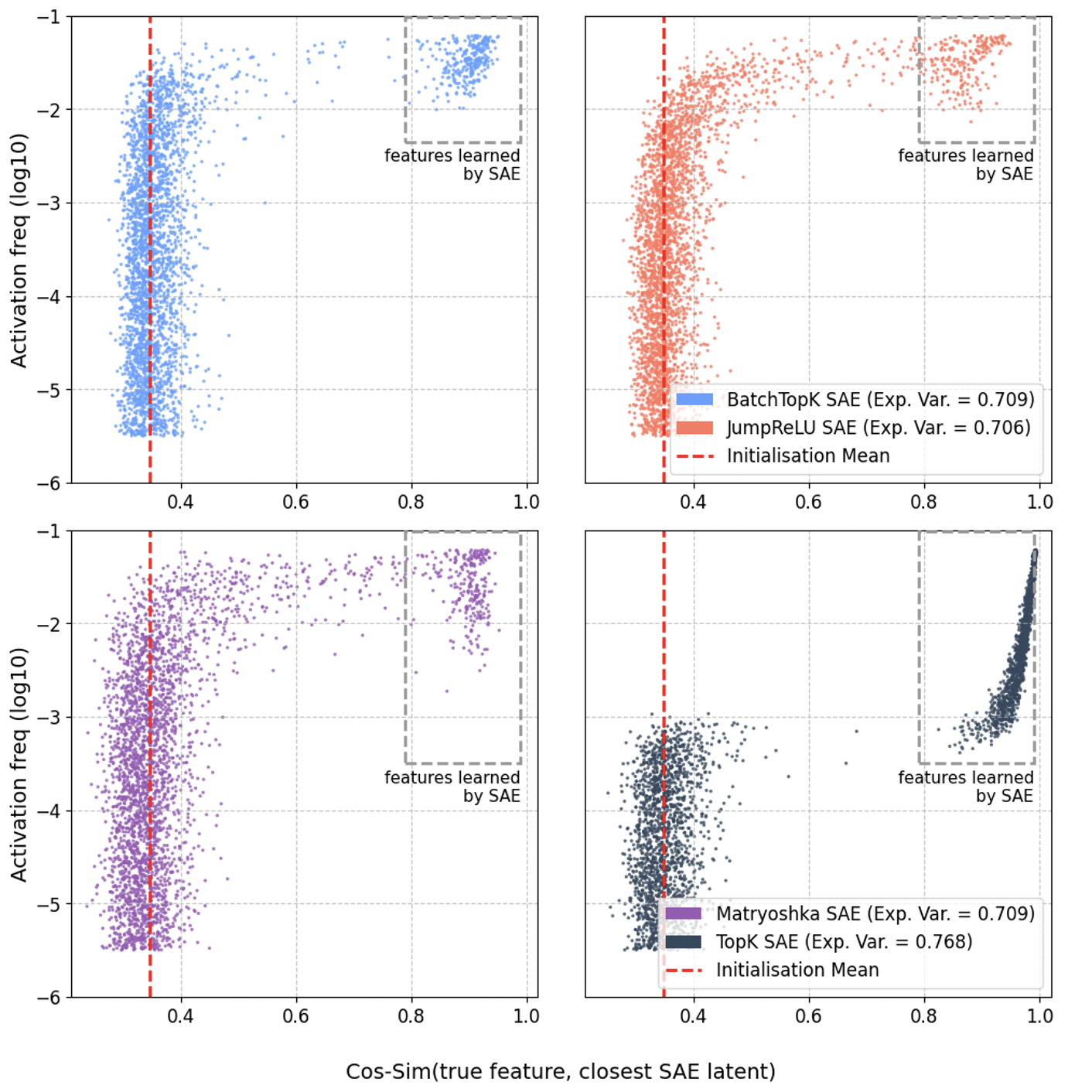}
\end{subfigure}
\caption{\textbf{SAEs performance on constant (Left) and variable (Right) probability settings}. BatchTopK, JumpReLU, and Matryoshka SAEs achieve high reconstruction fidelity (Explained Variance = 0.67--0.73) yet recover almost none of the ground-truth features, while TopK SAE achieves near-perfect recovery in the constant setting. In the variable setting, all architectures capture only the highest-frequency features, with TopK SAE again performing best.}
\label{fig:const}
\label{fig:vary}
\end{figure}

Our experimental results, summarized in Figure~\ref{fig:const}, reveal that SAEs fail at their core objective despite a strong reconstruction metric.



\textbf{Three out of four SAEs fail even in the simplest toy setting with fully aligned hyperparameters.} In the constant probability setting (Figure~\ref{fig:const}, left), BatchTopK, JumpReLU, and Matryoshka SAEs achieve an explained variance of approximately 0.67 - a value that would typically be interpreted as reasonably strong reconstruction. Yet they recover almost none of the ground‑truth features: for BatchTopK, only 3 out of 3200 true features have a cosine similarity above 0.8. Surprisingly, TopK SAE is the exception, achieving near‑perfect recovery (99.9\%). We attribute this to its stronger inductive bias: TopK enforces a hard sparsity constraint, leaving less room for optimization to find alternative representations compared to the softer constraints used in other architectures. Thus, for the three failing architectures, high reconstruction fidelity does not translate into discovering true features - exposing a fundamental limitation of current SAE evaluation.

\textbf{All SAEs recover only the most highly activated ground-truth features in the variable setting.} In the more realistic heavy-tailed setting, all four architectures achieve a high explained variance of 0.71, yet feature recovery remains limited. TopK SAE performs best at 43\%, while JumpReLU, BatchTopK, and Matryoshka recover only 7-9\% of features. Crucially, Figure~\ref{fig:vary} (right) shows this recovery is highly non-uniform: SAEs capture almost exclusively the highest-frequency features, leaving the entire long tail unmatched. Matryoshka's hierarchical structure slightly redistributes recovery toward mid-frequency features, but at the cost of missing some of the highest-frequency ones. Thus, the reconstruction objective biases SAEs toward the most frequent features, causing poor recovery despite strong reconstruction - a disconnect we next confirm on real LLM activations.


\section{Case Study \#2: Validating SAEs on real LLMs activations}

Having established SAE limitations in idealized settings, we now evaluate them on real LLM activations, where ground truth is unknown. In prior work \citep{karvonen2025saebench}, strong performance on four established proxy metrics has been taken as evidence that SAEs learn true model features: reconstruction fidelity, latent interpretability, sparse probing, and causal editing. We directly test this assumption by introducing three baselines where key SAE components are randomly initialized and frozen, preventing any learning beyond initialization.  By comparing these baselines to fully-trained SAEs, we perform a crucial null test: if current metrics reliably certify meaningful feature learning, fully-trained SAEs should strongly outperform models with frozen random components.


\subsection{Experimental setup}

\textbf{SAE variants and baselines.} 
We examine three SAE architectures: the state-of-the-art BatchTopK SAE \citep{bussmann2024batchtopk} and JumpReLU SAE \citep{rajamanoharan2024jumping}, as well as the traditional ReLU SAE \citep{bricken2023monosemanticity} (see results for TopK SAE in Appendix \ref{sec:topk_real}). To test if SAEs decompose model internal mechanisms rather than exploiting spurious correlations in data, we compare them against three baselines where key components are randomly initialized and frozen:

\begin{enumerate}
    \item \textbf{Frozen Decoder}: Decoder vectors $\mathbf{W}^{\text{dec}}_j$ are randomly initialized and frozen throughout training. This tests how well SAEs can perform when the latents are fixed to random vectors.

    \item \textbf{Soft-Frozen Decoder}: Decoder vectors $\mathbf{W}^{\text{dec}}_j$ are randomly initialized and constrained via projection to remain within a cosine similarity $\tau$ of their initial values throughout training (after each step, decoder columns outside the $\tau$ cap are projected back). This baseline is motivated by our observation that SAE loss plateaus early while decoder vectors remain near initialization. We hypothesize this indicates SAEs may operate in a \textit{lazy training} regime~\citep{chizat2019lazy, kumar2023grokking}, where reconstruction loss is dramatically reduced via small adjustments to latents without substantially changing their core semantics (see further discussion in Appendix~\ref{ref:diss}). This baseline directly tests that hypothesis. For all experiments, we use $\tau$ = 0.8 (see ablations on $\tau$ in Appendix~\ref{sec:tau_ablation}).

      \item \textbf{Frozen Encoder}: Encoder vectors $\mathbf{W}^{\text{enc}}_j$ are randomly initialized and frozen. This evaluates SAE performance when each feature's activation pattern (i.e., the contexts that trigger it) is predetermined in initialization, and only the activation threshold and decoder are learned.
\end{enumerate}

\textbf{Models and Configuration.}
Following \citet{bussmann2025learning}, we train all SAEs on the residual stream activations from layer 12 of the Gemma-2-2B model \citep{team2024gemma} (26 layers total). To assess generalizability, we also conduct experiments on layer 19 of Gemma-2-2B and layer 16 of Llama-3-8B \citep{dubey2024llama} (32 layers total). Each SAE uses a standard expansion factor $k$ of 32 \citep{bussmann2025learning} (matching our toy model experiments), resulting in dictionary sizes of 73,728 and 131,072 latents for Gemma-2-2B and Llama-3-8B, respectively. We sweep sparsity levels $\text{L}0$ in the set $\{80, 115, 160, 225, 320\}$ for each architecture.

All models are trained on 500 million tokens from the OpenWebText corpus \citep{Gokaslan2019OpenWeb} with a context length of 512. We use the AdamW optimizer \citep{loshchilov2017decoupled} with a learning rate of $2 \times 10^{-4}$ and a batch size of 4098. Training uses the open-source \texttt{saprmarks/dictionary\_learning} library. For \textit{cov} initialization, covariance is estimated from 100k OpenWebText tokens, vectors are sampled from $\mathcal{N}(0,\Sigma)$ and $\ell_2$-normalized; \textit{iso} samples uniformly from the unit sphere. To ensure a fair comparison, all models share identical initialization, data ordering, and hyperparameters. All training used a single NVIDIA A100-80GB GPU, with each model requiring $\approx$ 12 hours on 500M tokens. Main L0-sweep plots are single-seed; mean$\pm$std over 3 seeds at L0$=$160 appear in Appendix~\ref{sec:ablation}. We will publicly release all code, hyperparameters, and trained SAEs to ensure full reproducibility; the public repository will be available at \url{https://github.com/AntonKorznikov/sanity-checks-saes}.
Extended results for all layers and models (Gemma-2-2B layers 12 and 19, and Llama-3-8B layer 16) with detailed metrics and standard errors are provided in Appendix~\ref{sec:ablation}. A minimal evaluation protocol for new SAE architectures is in Appendix~\ref{sec:protocol}.

\textbf{Evaluation.} We evaluate the trained SAEs and baselines along four key dimensions: (1) reconstruction fidelity measured by Explained Variance \citep{gao2024scaling, rajamanoharan2024jumping}; (2) latent's interpretability evaluated through both automated scoring (AutoInterp) \citep{paulo2024automatically} and qualitative manual analysis of activation patterns; (3) sparse probing using the SAEBench framework \citep{karvonen2025saebench} to measure how precisely individual latents correspond to meaningful concepts (e.g., sentiment, syntax, factual recall); and (4) causal editing evaluated via the RAVEL framework \citep{huang2024ravel, karvonen2025saebench} to test whether SAE features can be used for precise and targeted model edits. 


\subsection{Results}

\begin{figure}[t]
  \centering
  \begin{subfigure}[b]{0.48\linewidth}
    \centering
    \includegraphics[width=\linewidth]{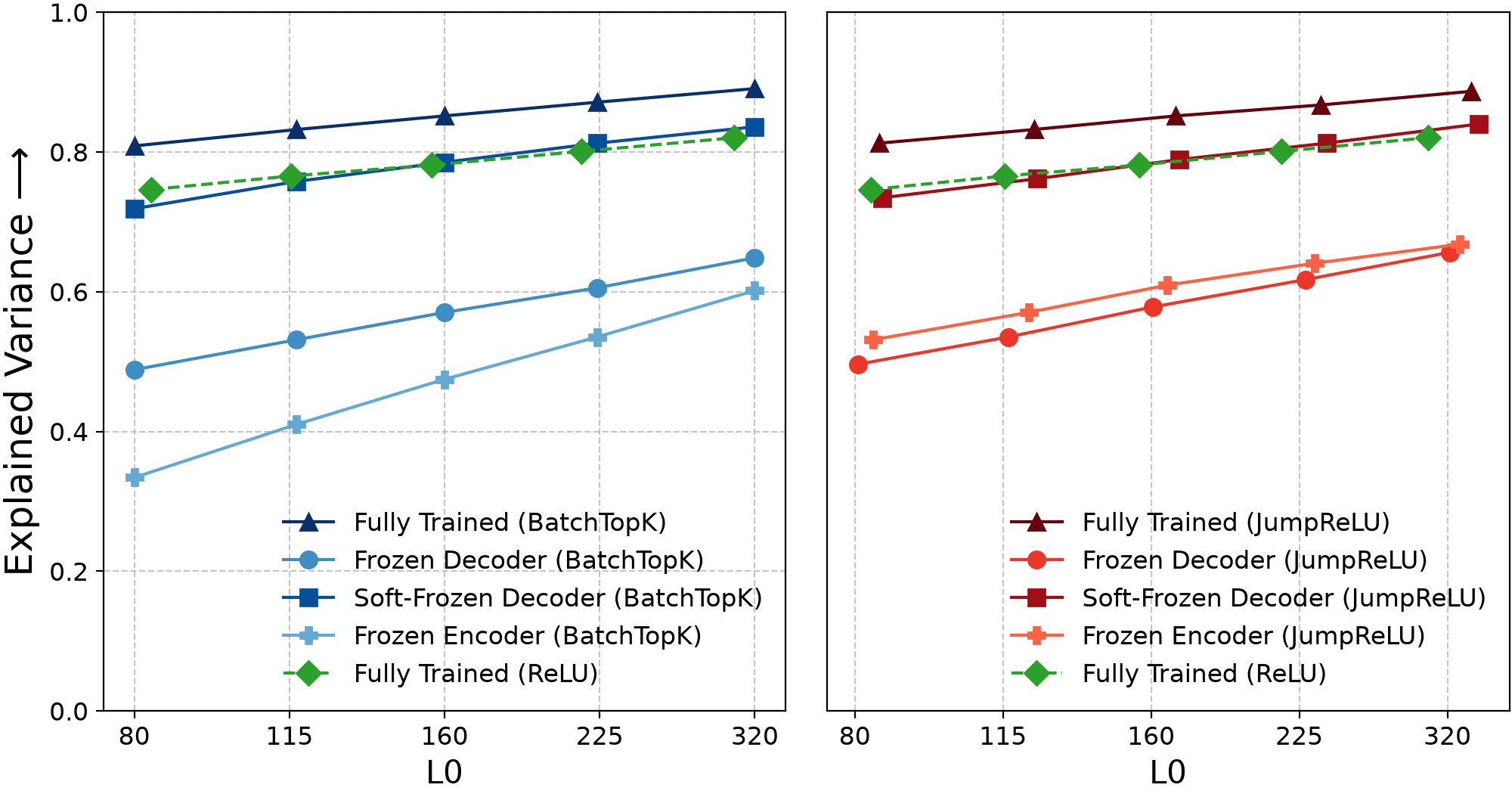}
  \end{subfigure}
  \hfill
  \begin{subfigure}[b]{0.48\linewidth}
    \centering
    \includegraphics[width=\linewidth]{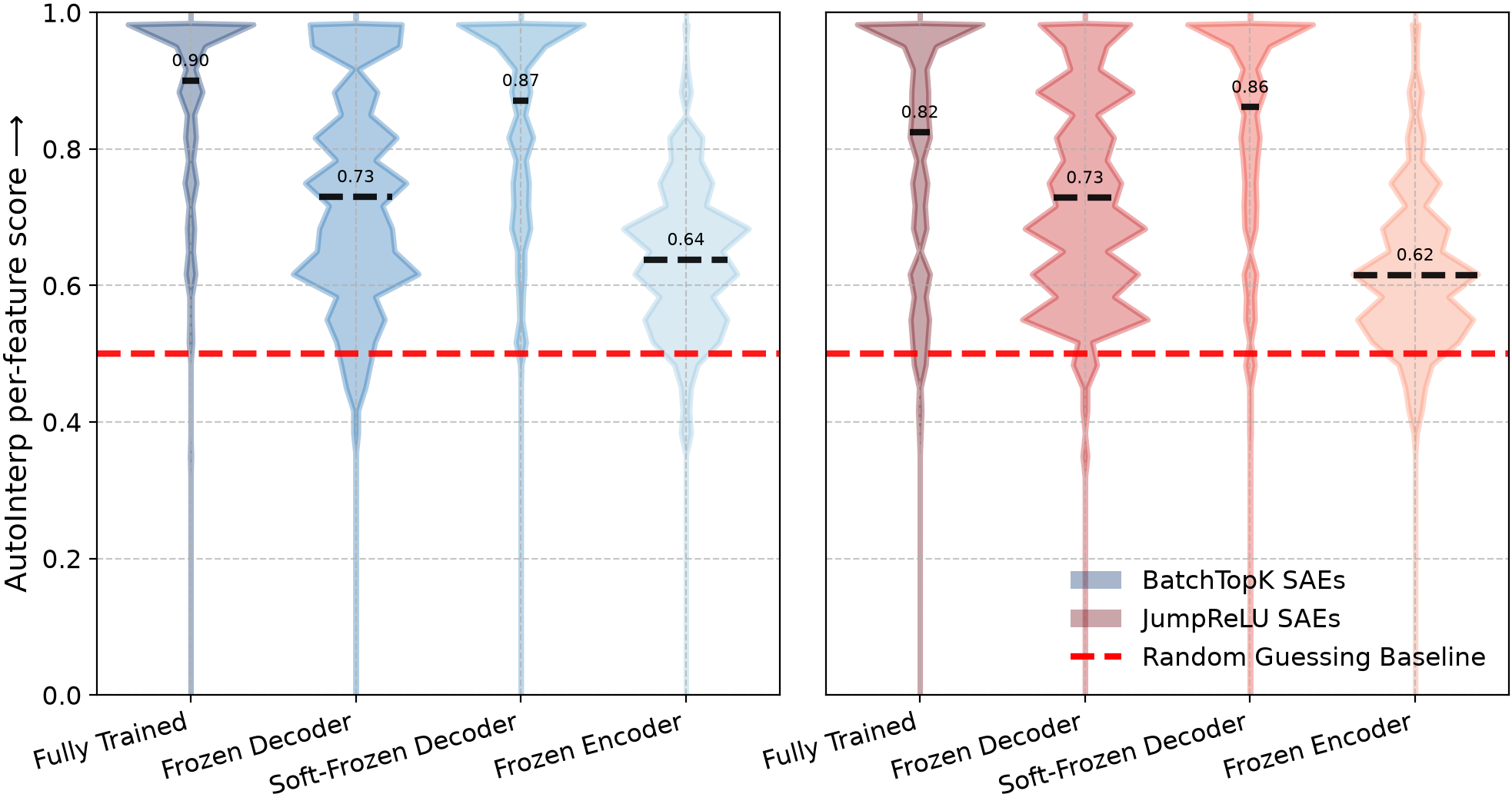}
  \end{subfigure}
  \caption{\textbf{Frozen SAE baselines on reconstruction (Left) and interpretability (Right).} Soft-Frozen SAEs match fully-trained ReLU explained variance and lose only 6\% relative to their original variants. AutoInterp scores stay high for both architectures, and the Soft-Frozen variant matches original performance.}
  \label{fig:var}
  \label{fig:autointerp}
\end{figure}

\begin{figure}[t]
    \centering
    \includegraphics[width=\linewidth]{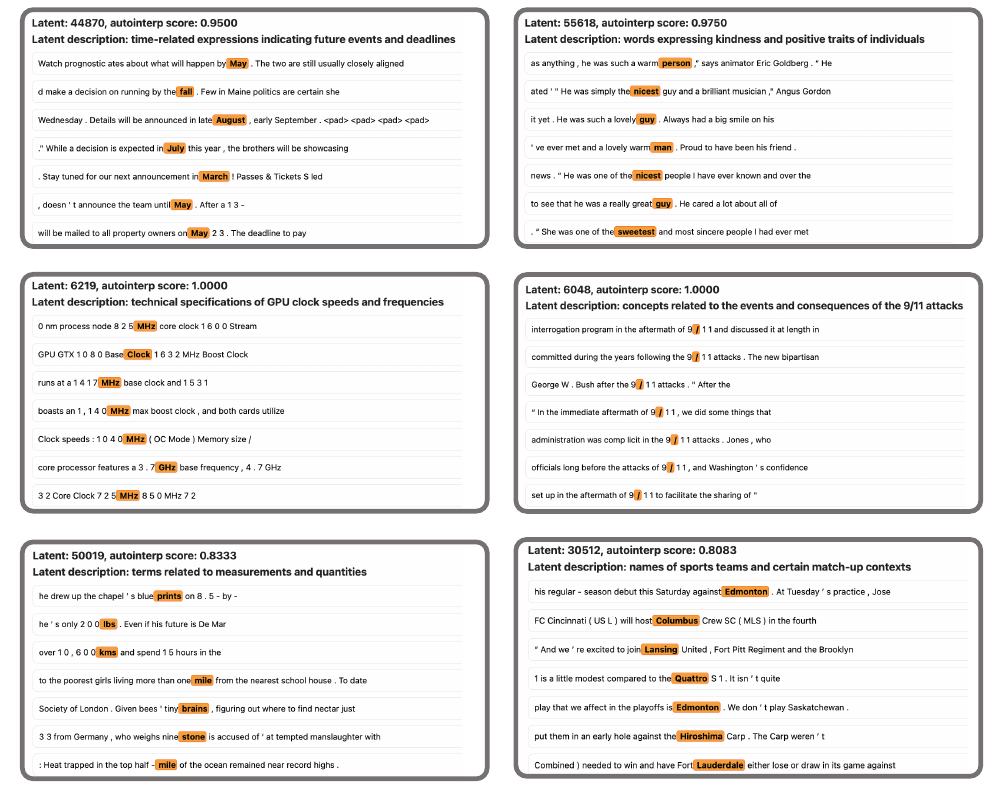}
    \caption{\textbf{Qualitative analysis: frozen SAEs produce interpretable latents.} Sample activation contexts for latents from Soft-Frozen Decoder (rows 1-2) and Frozen Encoder (row 3) variants of BatchTopK SAE, with corresponding descriptions and AutoInterp scores. The features capture abstract concepts, showing that high-level interpretability can emerge without learned feature alignment.}
    \label{fig:examples}
\end{figure}

\begin{figure}[t]
  \centering
  \begin{subfigure}[b]{0.47\linewidth}
    \centering
    \includegraphics[width=\linewidth]{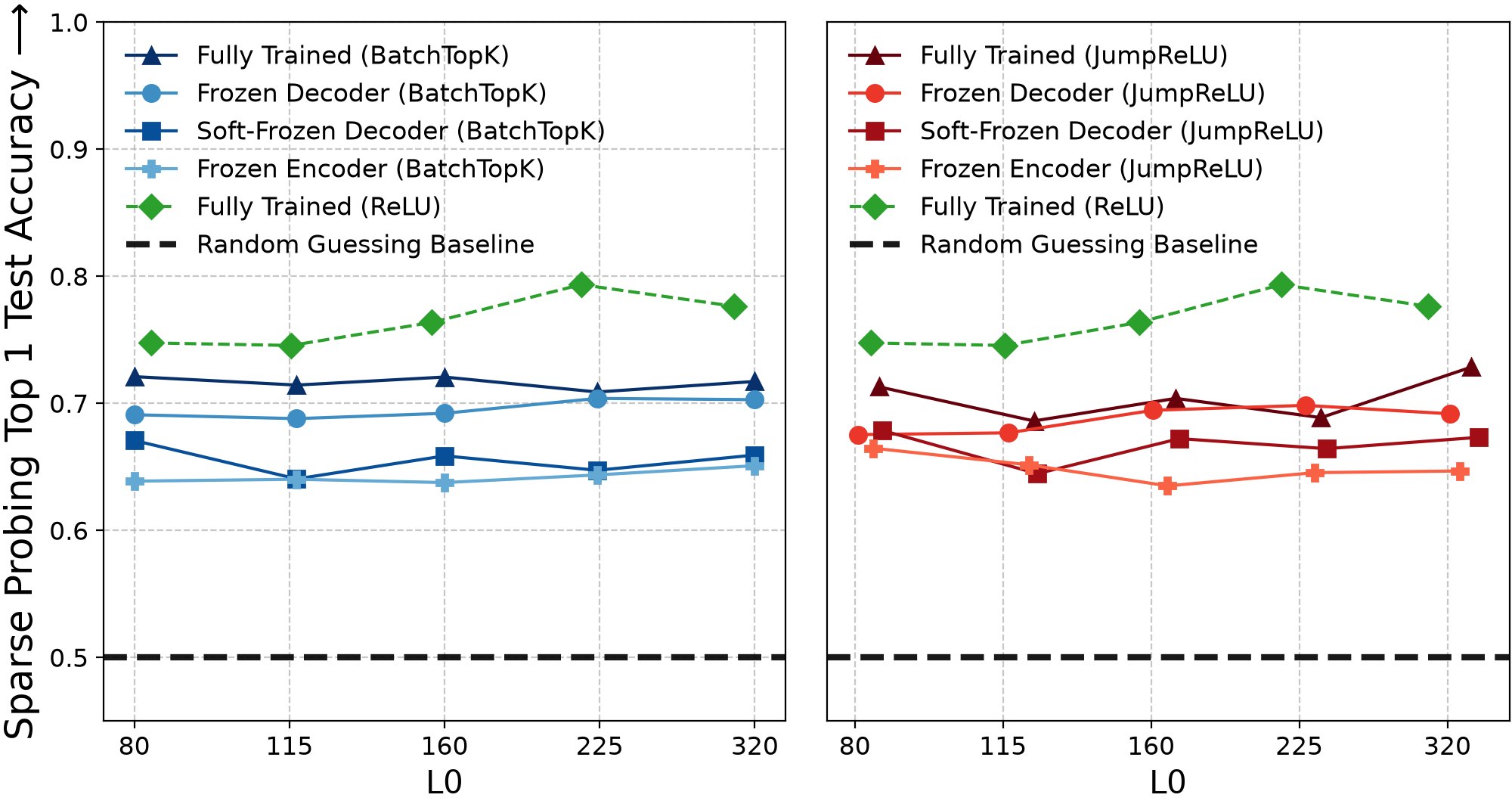}
  \end{subfigure}
  \hfill
  \begin{subfigure}[b]{0.47\linewidth}
    \centering
    \includegraphics[width=\linewidth]{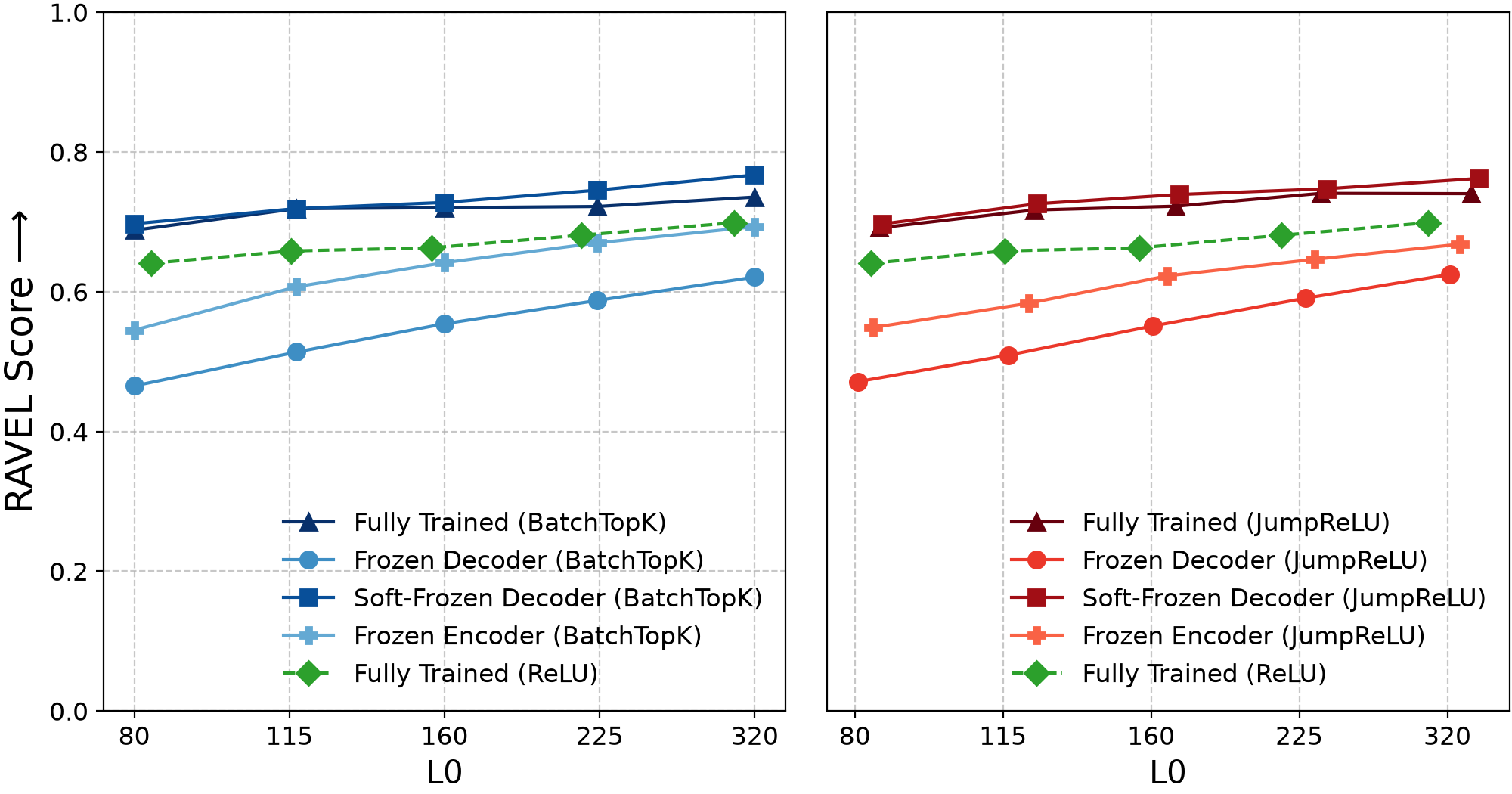}
  \end{subfigure}
  \caption{\textbf{Frozen SAE baselines on sparse probing (Left) and causal editing (Right).} Frozen BatchTopK and JumpReLU variants reach single-top-latent probing accuracy comparable to fully-trained SAEs. RAVEL disentanglement scores with frozen components are equivalent to fully-trained SAEs.}
  \label{fig:probing}
  \label{fig:ravel}
\end{figure}

Our results, summarized in Figures \ref{fig:var} to \ref{fig:ravel}, show that SAEs with frozen random components perform comparably to fully-trained SAEs across all four evaluation dimensions, suggesting that passing standard benchmarks may be necessary but is not sufficient evidence that SAEs learn meaningful feature decompositions.

\textbf{Frozen SAE baselines show strong reconstruction performance.} A fundamental premise of SAE evaluation is that strong reconstruction performance indicates the model has learned a meaningful decomposition of activations \citep{bricken2023monosemanticity}. We directly test this premise by evaluating whether SAEs maintain high reconstruction fidelity even when key components are fixed to random values. Following prior work \citep{gao2024scaling, rajamanoharan2024jumping}, we use Explained Variance (Eq.~\ref{eq:1}) as our primary reconstruction metric. We also measure cross-entropy loss and KL-divergence when substituting original activations with their SAE approximations, with these results provided in Appendix~\ref{sec:add}.

As shown in Figure~\ref{fig:var}, left, frozen SAE baselines achieve robust performance across both SAE architectures, with some baselines even matching the reconstruction fidelity of fully-trained SAEs. For instance, at L0=160, the original JumpReLU SAE attains an Explained Variance of 0.85, while its Soft-Frozen Decoder variant reaches 0.79. The Frozen Decoder and Frozen Encoder variants at the same sparsity yield scores of 0.58 and 0.60, respectively, providing that even severely constrained models can achieve non-trivial performance. This pattern indicates that SAEs can produce high-quality reconstructions even when their latent representations or activation patterns are randomized and fixed. Combined with our toy model results (Sec. \ref{toy_model_res}), this suggests that strong reconstruction performance may not be a sufficient indicator for validating the discovery of true underlying features.

\textbf{Latents from frozen SAE baselines are monosemantic and interpretable.} Beyond reconstruction, SAEs are claimed to produce interpretable latents, taken as proof they capture genuine model features \citep{bricken2023monosemanticity}. We test whether this interpretability emerges from meaningful feature learning or can be produced even when core components are frozen at random values.

To evaluate the interpretability of SAE latents, we follow the automated interpretability methodology outlined in \citep{paulo2024automatically, karvonen2025saebench}, leveraging LLMs to generate and validate human-readable feature descriptions. First, we select 200 random SAE latents for all SAE architectures and baseline variants with L0=160, excluding non-active ``dead'' features, defined as those with activation frequencies $<$ 1e-6. Dead-feature rates are negligible (fully-trained $0.1\%$, Frozen Decoder $0.5\%$, Frozen Encoder $1\%$); AutoInterp uses live latents only. For each latent, we collect up to 15 top-activating sequences and prompt GPT-4o-mini \citep{hurst2024gpt} to generate a concise description capturing the feature's core concept (e.g., \textit{``sentiment terms''} or \textit{``math expressions''}). To assess these descriptions, we create a test set for each latent consisting of 100 sequences: 50 sequences that activate the latent at varying strengths and 50 random non-activating sequences. A separate GPT-4o-mini model then predicts, based on the description, whether each sequence activates the latent, treating this as a binary classification task. The AutoInterp score is defined as the classification accuracy on this test set.

The AutoInterp scores in Figure~\ref{fig:autointerp}, right, reveal that frozen SAE variants produce monosemantic, interpretable latents nearly as well as their fully-trained counterparts. For instance, the Soft‑Frozen BatchTopK SAE attains a mean AutoInterp score of 0.88, closely matching the fully‑trained variant's score of 0.90. This suggests that high interpretability can emerge even when latent directions are constrained to remain near random initialization. Notably, the Frozen Encoder BatchTopK baseline, which fixes each feature's activation pattern to a random projection, still yields a modest fraction of highly interpretable latents, with roughly 30\% achieving AutoInterp scores above 0.7. 

To further validate these findings, we conducted a brief qualitative analysis. Focusing on the top-performing latents from our set of 200 randomly selected SAE features, we studied the sentences that triggered them (detailed activation contexts and descriptions for all evaluated latents are provided in the supplementary materials). We observed that many of these latents correspond to abstract, monosemantic, high-level concepts. As illustrated in Figure~\ref{fig:examples} (with examples from both Soft‑Frozen Decoder and Frozen Encoder variants), even our easy-to-implement baselines learn latents representing concepts such as \textit{``time-related expressions indicating future events and deadlines''}, \textit{``terms related to measurements and quantities''}, or \textit{``words expressing kindness and positive traits of individuals''}.

These results demonstrate that SAEs with frozen components achieve interpretability scores comparable to fully-trained SAEs, challenging the notion that high interpretability reflects meaningful feature learning. 


\textbf{Frozen SAEs yield competitive sparse probing results.} Sparse Probing evaluates the ability of SAEs to isolate specific concepts, such as sentiment, within individual latents without explicit supervision. For each concept, we select the top-\textit{k} latents by comparing their mean activations on positive versus negative examples. A linear probe is then trained on these latents to predict the concept. High probe accuracy indicates that the latents effectively capture the target concept in a disentangled manner. Following \citet{karvonen2025saebench}, we evaluate across five diverse concept types: profession classification, product classification and sentiment analysis, language identification, programming language classification, and news topic categorization.

As shown in Figure~\ref{fig:probing}, left, sparse probing accuracy remains strong even when core SAE components are randomly initialized and frozen. Under the single-top-latent ($k$=1) setup, frozen variants of both BatchTopK and JumpReLU SAEs achieve accuracy scores comparable to their fully-trained counterparts across all sparsity levels (for $k$=5 see Appendix C). For instance, at L0=225, the BatchTopK Frozen Decoder variant reaches 0.70, matching the fully-trained model's score, while the Frozen Encoder variant attains 0.65, a difference of only 0.05. These gaps remain narrow (0.001 to 0.07) across all settings and stay well above the random-guessing baseline of 0.50.

Together, these findings indicate that sparse probing scores can be 
inflated by dictionary scale alone: with tens of thousands of latents, 
some will inevitably correlate with concept clusters by chance, 
limiting its value as a standalone indicator of meaningful feature 
learning. We further discuss whether SAEs may learn features 
collectively rather than individually in 
Appendix~\ref{sec:atomic_vs_aggregate}.



\textbf{Frozen SAEs retain strong RAVEL causal editing performance.} To further test if SAEs discover genuinely meaningful model features, we examine their causal properties. We evaluate whether SAE latents encode causally independent concepts using the RAVEL framework \citep{huang2024ravel, karvonen2025saebench}, which tests if targeted interventions on latents can cleanly edit model predictions. This framework assesses whether the apparent causal utility of SAEs depends on learned feature alignment or can be achieved with frozen random components.

Given a factual statement like \textit{``Paris is in France''}, we encode the subject token (e.g., \textit{``Paris''}) with the SAE, train a binary mask to transfer latent values from a different token (e.g., \textit{``Tokyo''}), and decode the modified latents back into the residual stream. The model then generates completions; a successful edit would change the targeted attribute (e.g., now stating \textit{``Paris is in Japan''}) while preserving other attributes (e.g., still \textit{``People in Paris speak French''}). The RAVEL disentanglement score averages cause (target attribute change) and isolation (other attributes unchanged).


As shown in Figure~\ref{fig:ravel}, right, both BatchTopK and JumpReLU SAE variants with frozen components achieve RAVEL scores competitive with their fully-trained counterparts. For BatchTopK, the fully trained model achieves a RAVEL score of approximately 0.72–0.74 across higher sparsity levels (L0 $\geq$ 160). The Frozen Decoder variant performs surprisingly well, reaching 0.57-0.62, demonstrating that effective causal editing can be achieved even when decoder vectors are frozen as random directions. The Frozen Encoder variant also shows strong performance, maintaining a score of 0.63 for L0 = 160. Meanwhile, the Soft-Frozen Decoder variant matches or even exceeds the fully trained model, reaching up to 0.78 at L0 = 320. 

Thus, causal editing performance alone cannot distinguish learned representations from random vector coverage, revealing a gap in the discriminative power of current evaluation metrics.



\section{Limitations}
Our synthetic setup assumes independent feature activations, omitting correlations present in real networks. Since three out of four architectures fail even in this simplified sanity check, a more realistic setting would only increase the difficulty. How to incorporate structured covariances without arbitrary assumptions remains an open question.
Structured-correlation synthetic results on SynthSAEBench~\citep{chanin2026synthsaebench} are reported in Appendix~\ref{sec:synthsaebench}.
We focus on standard SAE architectures and do not evaluate related approaches such as transcoders or crosscoders~\citep{dunefsky2024transcoders}; the same frozen-encoder/decoder protocol can be applied to them, but we have not run those experiments and do not claim numerical gaps transfer.



\section{Discussion and Conclusion}

We evaluated SAEs through two complementary approaches: synthetic experiments with known ground-truth features, and comparisons against random baselines on real LLM activations. In synthetic settings with fully aligned hyperparameters, three out of four SAE architectures achieve 71\% explained variance yet recover only 7–9\% of true features - a failure mode invisible to reconstruction metrics alone. On real LLM activations, our frozen baselines match fully‑trained SAEs across reconstruction, interpretability, sparse probing, and causal editing, showing that passing standard benchmarks is not sufficient evidence of meaningful feature decomposition.

The disconnect between reconstruction and feature recovery is expected from classical sparse dictionary learning (NP-hard / non-identifiable under weak conditions)~\citep{tillmann2014computational,razaviyayn2015computational,spielman2012exact};
our synthetic results empirically confirm that good reconstruction need not recover ground-truth features.
When decoder directions remain near-random, we hypothesize that the encoder mainly performs amortized sparse selection over an expressive random dictionary; we do not claim a confirmed mechanism.
Our baselines are simple to implement and serve as minimal sanity checks: if a new SAE architecture substantially outperforms them, that provides stronger evidence for genuine feature learning (Appendix~\ref{sec:protocol}). We emphasize that our findings are a critique of current evaluation methodology, not a verdict against SAEs as a paradigm - and we hope they motivate more rigorous evaluation standards in mechanistic interpretability.

\begin{ack}
The authors received no specific funding for this work beyond their institutional affiliations.
\subsection*{Competing interests}
The authors declare no competing interests.
\end{ack}

\nocite{*}
\bibliographystyle{abbrvnat}
\bibliography{refs}


\newpage
\appendix

\section{SAEs: Claimed Benefits vs. Documented Limitations}
\label{app:sae_overview}

Table \ref{tab:sae_pros_cons} provides a comprehensive overview of both the claimed advantages that have motivated widespread adoption of SAEs and the documented challenges that cast doubt on their reliability.

\begin{table}[ht!]
\centering
\caption{\textbf{Sparse Autoencoders: Claimed Benefits vs. Documented Limitations}}
\small
\begin{tabular}{p{0.46\textwidth}|p{0.46\textwidth}}
\toprule
\textbf{Claimed Benefits} & \textbf{Documented Limitations} \\
\midrule
\textbf{Core Purpose:} Decompose dense activations into sparse, interpretable, monosemantic features \citep{bricken2023monosemanticity, cunningham2023sparse} &
\textbf{Spurious Features:} Find features even in randomly initialized transformers with no learned structure \citep{heap2025sparse} \\
\textbf{Monosemanticity:} Address polysemanticity by learning features that correspond to single coherent concepts \citep{bricken2023monosemanticity, cunningham2023sparse} &
\textbf{Unfaithfulness:} Reconstructed features do not faithfully represent true model computations \citep{leask2025sparse, menon2025analyzing} \\
\textbf{Scalability:} Successfully scaled to frontier models including Claude 3 Sonnet, GPT-4, and Gemma 2 \citep{templeton2024scaling, gao2024scaling, lieberum2024gemma} &
\textbf{Poor Generalization:} Fail to generalize across tasks, settings, and perturbations \citep{heindrich2025sparse, kantamneni2025sparse, li2025interpretability} \\
\textbf{AI Safety Applications:} Enable mechanistic interpretability for understanding safety mechanisms, reasoning, and misalignment \citep{wang2022interpretability, galichin2025have} &
\textbf{Downstream Performance:} Underperform on downstream tasks; other approaches often outperform SAEs \citep{wu2025axbench, smithnegative} \\
\textbf{Reconstruction Quality:} Achieve high reconstruction fidelity \citep{bricken2023monosemanticity, gao2024scaling} &
\textbf{Training Fragility:} Highly sensitive to hyperparameters and initialization \citep{chanin2025sparse, paulo2025sparseautoencoderstraineddata, minegishi2025rethinking} \\
\bottomrule
\end{tabular}
\label{tab:sae_pros_cons}
\end{table}

\section{Frozen Baselines for TopK SAE on Real Activations}
\label{sec:topk_real}

\begin{figure}[t]
\centering
\includegraphics[width=0.8\textwidth]{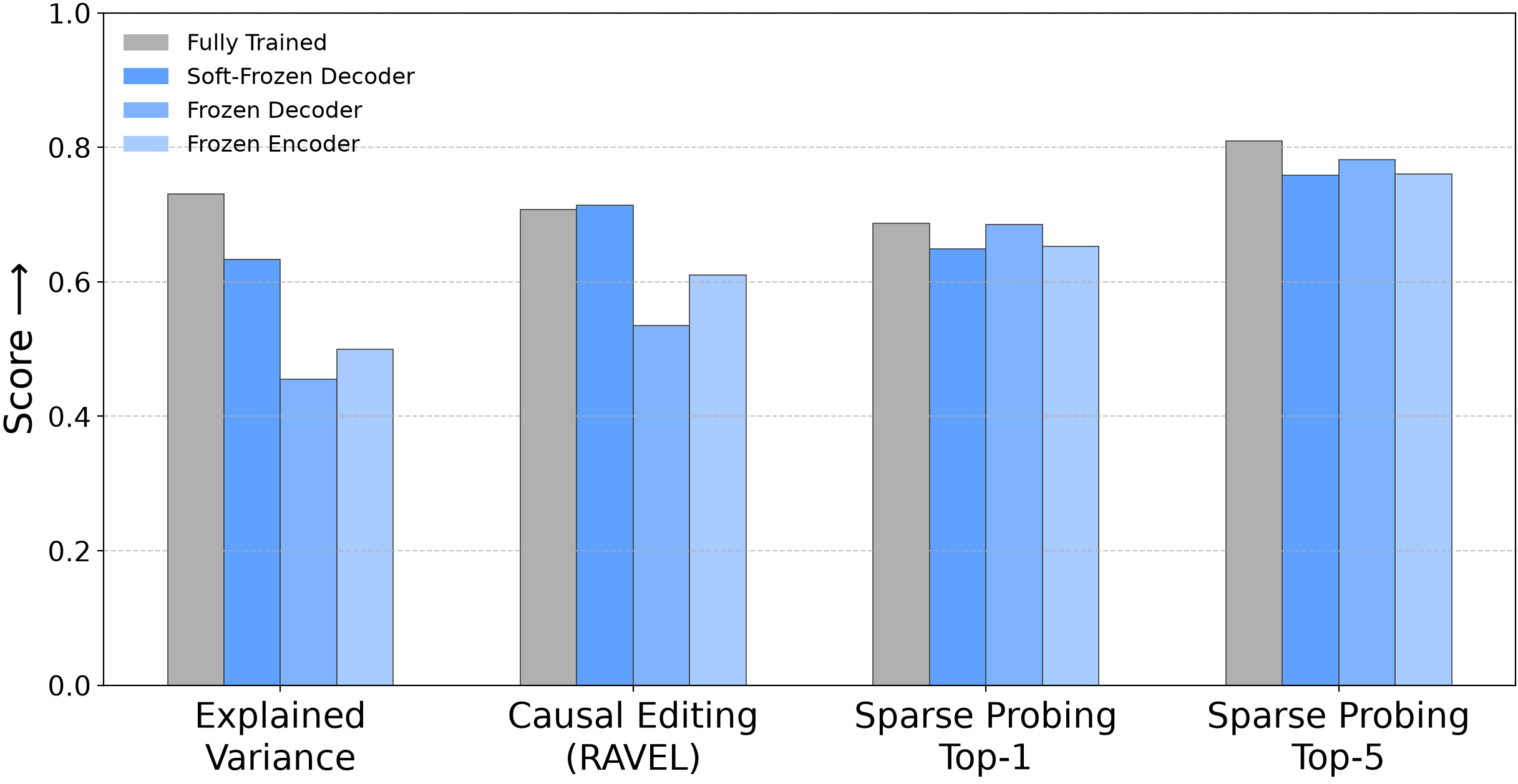}
\caption{\textbf{Frozen baseline performance for TopK SAE (L0=160) on Gemma-2-2B layer 12.}}
\label{fig:topk_frozen}
\end{figure}

To provide a comprehensive evaluation across SAE architectures, we extend our frozen baseline analysis to the simpler TopK SAE variant. Figure~\ref{fig:topk_frozen} presents the performance of TopK SAE (L0=160) trained on Gemma-2-2B layer 12 activations, comparing the four variants (Fully Trained, Soft-Frozen Decoder, Frozen Decoder, Frozen Encoder) across our standard evaluation metrics.

Notably, despite TopK SAE's strong performance in the synthetic setting 
(Section~\ref{sec:toy}), this advantage does not transfer to real LLM activations: frozen variants of TopK SAE remain competitive with the 
fully trained model, mirroring the pattern observed for BatchTopK and 
JumpReLU. This suggests that TopK's superior feature recovery in 
controlled synthetic settings is an artifact of that simplified setup 
rather than a general property of the architecture.

\section{The Soft-Frozen Decoder: Testing the Lazy Training Hypothesis}
\label{ref:diss}

The Soft-Frozen Decoder baseline directly tests whether SAEs operate in what we term a \textit{lazy training} regime~\citep{chizat2019lazy, kumar2023grokking}. Under this hypothesis, SAEs achieve strong reconstruction performance primarily through vanishingly small adjustments to both encoder and decoder vectors that remain semantically close to their random initializations without discovering fundamentally new feature directions. This baseline constrains decoder vectors to maintain at least $\tau$ = 0.8 cosine similarity with their initial random directions throughout training. Below, we present empirical observations and theoretical analysis that motivate this baseline and reveal its implications.

\begin{figure}[t]
\centering
\begin{subfigure}[t]{0.49\linewidth}
\centering
\includegraphics[width=\linewidth]{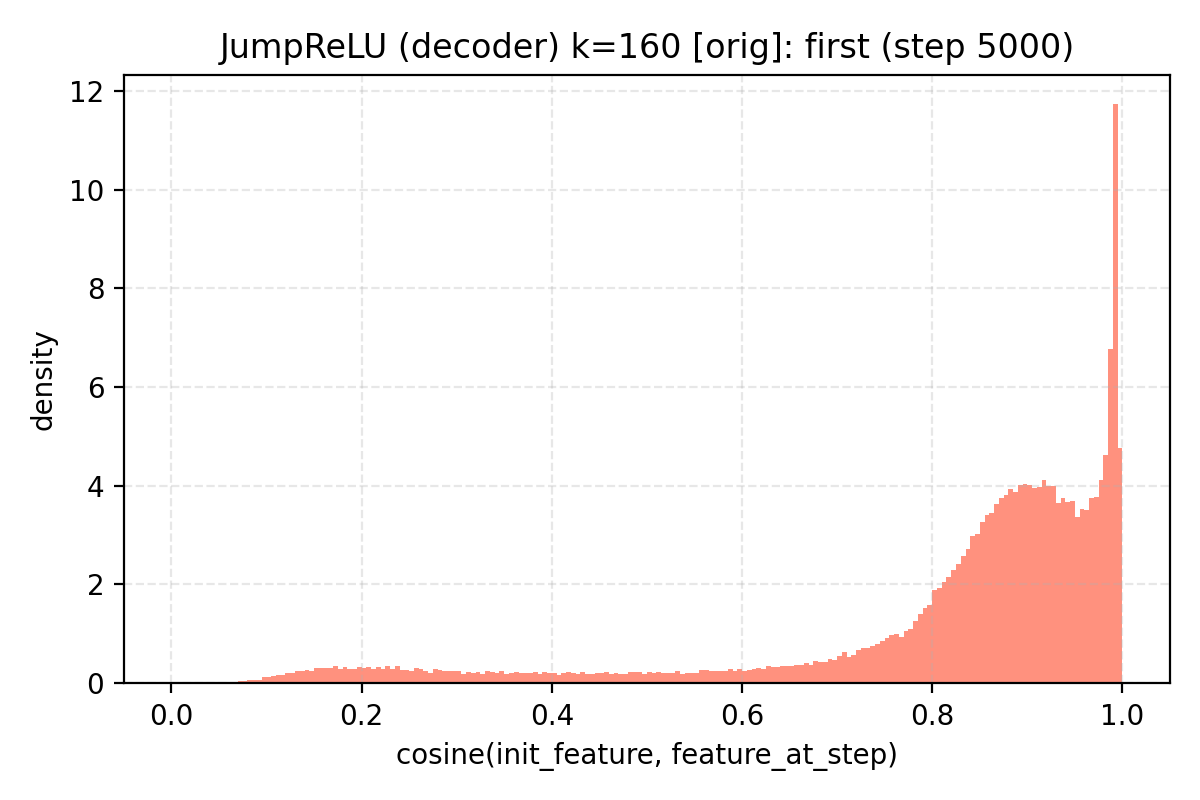}
\end{subfigure}
\hfill
\begin{subfigure}[t]{0.49\linewidth}
\centering
\includegraphics[width=\linewidth]{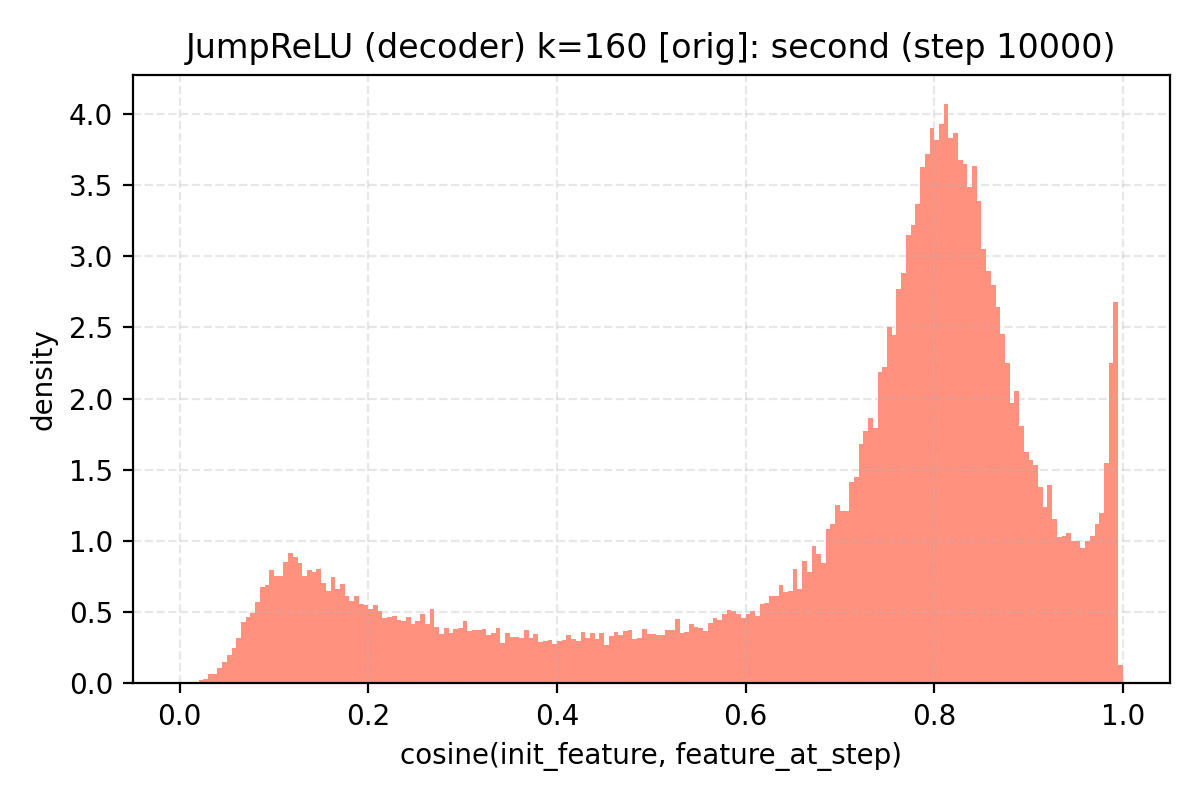}
\end{subfigure}
\caption{\textbf{Decoder vectors remain near random initialization throughout training.} Histograms of cosine similarity between decoder vectors at initialization and after 5\% (left) and 10\% (right) of training steps for a JumpReLU SAE. The strong concentration near 1.0 indicates minimal directional change even after reconstruction performance has plateaued, supporting the lazy training hypothesis and motivating the Soft-Frozen Decoder baseline.}
\label{fig:cosine_hist}
\end{figure}

\paragraph{Empirical Motivation: Early Convergence with Minimal Directional Change}
During training of a JumpReLU SAE (L0=160), we observed that the loss and explained variance plateaued early, while the decoder vectors remained close to their initial random directions. After just 5\% of total training steps, when loss and Explained Variance had plateaued at 0.80 (versus a final value of 0.86), the majority of decoder vectors showed minimal directional deviation from initialization (Figure~\ref{fig:cosine_hist}, left). This pattern persisted at 10\% training (Explained Variance = 0.82), with cosine similarities still concentrated near 0.8 (Figure~\ref{fig:cosine_hist}, right). These observations suggest SAEs achieve most reconstruction gains through vanishingly small adjustments to both encoder and decoder vectors rather than discovering new directional representations. Motivated by this finding and the theoretical analysis below, we set $\tau = 0.8$ for the Soft-Frozen Decoder baseline, selecting a threshold that permits very limited directional adjustment while remaining consistent with the observed training dynamics.

\paragraph{Theoretical Analysis: The Vanishing Probability of Feature Alignment}
The Soft-Frozen Decoder baseline initializes decoder vectors $\mathbf{W}^{\text{dec}}_i$ uniformly from the unit sphere $S^{n-1}$ and constrains them to maintain cosine similarity $\tau = 0.8$ with their initial values $\widehat{\mathbf{W}^{\text{dec}}_i}$. For the SAE to represent a specific ground-truth feature (e.g., \textit{``there is a knight on F3''} in a chess model~\citep{karvonen2024measuring}), at least one decoder vector must align closely with that feature's true direction.

Formally, the constraint confines each decoder vector to a spherical cap around its random initialization. The set of expressible directions is:
\begin{equation}
X := \{\mathbf{x} \in S^{n-1} \mid \exists i \in [m] \text{ such that } \langle \mathbf{x}, \widehat{\mathbf{W}}^{\text{dec}}_i \rangle \geq \tau\}.
\end{equation}

For an arbitrary ground-truth feature direction $\mathbf{Y} \in S^{n-1}$, the probability it lies within $X$ is bounded by:
\begin{equation}
P(\mathbf{Y} \in X) \leq m \cdot P(\langle \mathbf{Y}, \widehat{\mathbf{W}}^{\text{dec}}_0 \rangle \geq \tau).
\end{equation}

The probability that a random unit vector falls within a spherical cap of cosine similarity $t$ decays exponentially with dimension $n$~\citep{tkocz2012upper}. With our experimental parameters ($\tau=0.8$, $n=2304$, $m=73728$):
\begin{align}
P(\langle \mathbf{Y}, \widehat{\mathbf{W}}^{\text{dec}}_0 \rangle \geq \tau) &\leq \exp\left(-\frac{n \tau^2}{2}\right) \approx 6.36 \times 10^{-321}, \\
P(\mathbf{Y} \in X) &\leq 4.67 \times 10^{-316}.
\end{align}

Thus, the Soft-Frozen Decoder is mathematically unlikely to align with arbitrary semantic features beyond chance initialization. Its strong performance must therefore arise from efficiently combining its nearly-fixed, random directions through vanishingly small adjustments to both encoder and decoder vectors. This provides formal evidence that standard SAE metrics can be optimized without learning meaningful feature directions, challenging the assumption that reconstruction fidelity indicates successful feature discovery.

\paragraph{Discussion}
The competitive performance of Soft-Frozen Decoder SAEs across multiple evaluation dimensions (Figure~\ref{fig:main_figure}) provides strong support for the lazy training hypothesis: much of what we attribute to learned feature discovery may instead reflect efficient use of nearly-fixed, randomly initialized components with only minimal adjustments. This success challenges the premise that SAEs learn meaningful feature decompositions through training, and underscores the importance of rigorous validation against constrained baselines in interpretability research.

\section{Frozen Models Ablation}
\label{sec:ablation}

We provide an ablation study comparing fully-trained SAEs against our frozen baselines under two initialization schemes: \textit{iso} (vectors sampled uniformly from the unit sphere) and \textit{cov} (vectors sampled from a Gaussian distribution with zero mean and covariance estimated from real activations, then normalized). Tables~\ref{tab:layer12}, \ref{tab:layer19}, and \ref{tab:layer16} report key metrics (mean $\pm$ standard deviation over 3 seeds) at L0=160 for Gemma-2-2B layers 12 and 19, and Llama-3-8B layer 16. The experiments in the preceding sections use the best-performing initialization for each baseline: \textit{cov} for Frozen Decoder and \textit{iso} for Soft-Frozen Decoder and Frozen Encoder. These results confirm that the competitive performance of frozen baselines is consistent across layers and model families. One-sided Welch $t$-tests ($H_0{:}\,\mu(\text{fully-trained})\le\mu(\text{frozen})$; $n{=}3$ seeds) across frozen variants in these tables reach significance for Explained Variance in 30/30 comparisons, Sparse Probing (top-1) in 21/30, and Causal Editing in 20/30; for roughly a quarter of downstream comparisons the gap is not statistically significant. Stars in the tables mark $p{<}0.05/0.01/0.001$ ($^*$/$^{**}$/$^{***}$).

\section{Additional Metrics}
\label{sec:add}

To complement our primary evaluation metrics, we assess SAEs and frozen baselines using additional performance measures following the SAEBench framework \citep{karvonen2025saebench}, which evaluates cross-entropy loss and KL-divergence when substituting original activations with SAE reconstructions. As shown in Figures \ref{fig:ce} and \ref{fig:kl}, Frozen Decoder and Frozen Encoder baselines show modest degradation in these metrics compared to fully-trained SAEs. In contrast, the Soft-Frozen Decoder variant closely matches, and sometimes exceeds, the performance of fully-trained ReLU SAEs, demonstrating that even minimal adjustments to decoder vectors can achieve strong reconstruction quality. When moving from single‑latent to multi‑latent sparse probing (top‑5, Figure \ref{fig:top5}), frozen baselines show only modest gaps relative to fully‑trained SAEs (3–8\%, as detailed in Appendix~\ref{sec:atomic_vs_aggregate}), indicating that even collective feature structure does not strongly favor learned representations over random baselines.

\section{Atomic vs. Aggregate Feature Learning}
\label{sec:atomic_vs_aggregate}

A natural objection to our findings is that SAEs may not be intended 
to learn features in the strict atomic sense, where each individual 
decoder vector corresponds to a single, human-interpretable concept, 
but rather collectively, where groups of decoder vectors together span 
a meaningful semantic subspace. Under this weaker hypothesis, the 
relevant question is not whether individual latents are meaningful, 
but whether the learned dictionary as a whole provides a useful 
structured decomposition of activation space.

We note, however, that the atomic interpretation is precisely what 
motivates the primary use cases of SAEs in mechanistic interpretability: 
individual latents are labeled, steered, and causally intervened upon 
in isolation~\citep{bricken2023monosemanticity, templeton2024scaling}. 
A decomposition where no individual decoder vector corresponds to a 
semantic concept, but groups of them collectively encode useful 
structure, would undermine these applications, as any apparent 
interpretability of individual latents would be an artifact of 
evaluation rather than evidence of genuine feature discovery.

That said, our experiments provide evidence against even this weaker 
collective hypothesis. The top-5 sparse probing task, in which a 
logistic probe is trained on the five most concept-predictive latents, 
offers a partial test: it measures whether the learned dictionary 
contains at least a small cluster of latents that collectively encode 
a target concept. As shown in Tables~\ref{tab:layer12}--\ref{tab:layer16}, 
fully-trained SAEs do not substantially outperform frozen baselines 
even in this setting. The gaps between fully-trained SAEs and their 
best-performing frozen counterparts range from 3.0\% (BatchTopK on 
Llama-3-8B layer 16) to 7.9\% (BatchTopK on Gemma-2-2B layer 19), 
with frozen encoders, whose activation patterns are entirely random, 
remaining competitive across all settings. Thus, even when evaluated 
on collective rather than individual feature structure, frozen baselines 
offer only marginally weaker evidence of meaningful decomposition than 
fully-trained SAEs.

\section{Ablation on the Soft-Frozen Decoder Threshold}
\label{sec:tau_ablation}
 
The Soft-Frozen Decoder constrains decoder vectors to maintain cosine similarity $\tau \geq 0.8$ with their random initialization throughout training. While $\tau = 0.8$ is our primary setting (motivated in Appendix~\ref{ref:diss}), one might ask whether downstream metrics would degrade under stricter constraints. To address this, we ran an ablation with $\tau \in \{0.8, 0.9, 0.95\}$ for BatchTopK SAE (L0=160) on layer 12 of Gemma-2-2B.
 
As shown in Table~\ref{tab:tau_ablation}, tightening the constraint 
produces the expected degradation in Explained Variance, which drops 
from 0.785 to 0.691 as $\tau$ increases from 0.8 to 0.95 - a 
natural consequence of leaving less room for directional adjustment. 
However, the downstream task metrics tell a strikingly different story: 
Sparse Probing and Causal Editing scores remain largely stable across 
all constraint levels despite the reconstruction penalty. This 
asymmetry reinforces our main thesis: while reconstruction fidelity is 
sensitive to how tightly the decoder is constrained, the 
interpretability-oriented metrics that are typically used to validate 
SAEs are surprisingly robust to this constraint, and therefore 
insufficient to certify meaningful feature learning.
 
\begin{table}[t]
\caption{Evaluation metrics for Soft-Frozen Decoder across constraint values $\tau$.}
\centering
\resizebox{\textwidth}{!}{
\begin{tabular}{lccc}
\hline
\textbf{Method} & \textbf{Explained Variance} $\uparrow$ & \textbf{Sparse Probing (Top-1)} $\uparrow$ & \textbf{Causal Editing} $\uparrow$ \\
\hline
Fully-Trained SAE       & 0.852 & 0.721 & 0.720 \\
\hline
Soft-Frozen ($\tau = 0.80$) & 0.785 & 0.659 & 0.728 \\
Soft-Frozen ($\tau = 0.90$) & 0.734 & 0.639 & 0.712 \\
Soft-Frozen ($\tau = 0.95$) & 0.691 & 0.637 & 0.681 \\
\hline
\end{tabular}}
\label{tab:tau_ablation}
\end{table}

\section{Random SAEs on CLIP}
To validate our findings from language models, we conduct a parallel analysis on vision models by comparing trained and randomly initialized SAEs on CLIP ViT-B/32 \citep{radford2021learning}. We
  use pretrained SAEs \citep{joseph2025prismaopensourcetoolkit} from layers 3, 5, 7, and 9 and create matched random baselines with identical architectures but Kaiming-initialized weights. For each SAE, we select features with rare  ($<0.1 \%$)  activation frequency, then visualize images from ImageNet-10k \citep{deng2009imagenet} at different activation percentiles (100th, 75th, 50th, 25th) for each feature. This
  percentile-based visualization reveals whether features capture coherent semantic concepts. Some trained features (layers 3 and 9) show clearer monosemantic concepts, while many at layers 5 and 7 remain ambiguous, so the distinction from random baselines is not always sharp. We present these visualisations in Figures \ref{fig:clip_layer3_comparison}, \ref{fig:clip_layer5_comparison}, \ref{fig:clip_layer7_comparison} and \ref{fig:clip_layer9_comparison}

\newpage

\begin{table}[h]
\caption{Frozen SAEs performance on Gemma-2-2B layer 12 L0=160.}
\centering
\resizebox{\textwidth}{!}{%
\begin{tabular}{lcccc}
\hline
\textbf{Variant} & \textbf{Explained Variance} & \textbf{Sparse Probing (top-1)} & \textbf{Sparse Probing (top-5)} & \textbf{Causal Editing} \\
\hline
\textbf{BatchTopK:} \\
\quad Fully-Trained (iso) & 0.852 $\pm$ 0.001 & 0.721 $\pm$ 0.026 & 0.849 $\pm$ 0.025 & 0.720 $\pm$ 0.049 \\
\quad Frozen Decoder (iso) & 0.430 $\pm$ 0.000$^{***}$ & 0.669 $\pm$ 0.010$^{*}$ & 0.768 $\pm$ 0.013 & 0.562 $\pm$ 0.027$^{**}$ \\
\quad Frozen Decoder (cov) & 0.570 $\pm$ 0.004$^{***}$ & 0.692 $\pm$ 0.014 & 0.797 $\pm$ 0.014 & 0.554 $\pm$ 0.028$^{**}$ \\
\quad Soft-Frozen Decoder (iso) & 0.785 $\pm$ 0.001$^{***}$ & 0.659 $\pm$ 0.035$^{*}$ & 0.723 $\pm$ 0.031 & 0.728 $\pm$ 0.038 \\
\quad Frozen Encoder (iso) & 0.475 $\pm$ 0.004$^{***}$ & 0.638 $\pm$ 0.013$^{**}$ & 0.731 $\pm$ 0.013 & 0.641 $\pm$ 0.027$^{*}$ \\
\quad Frozen Encoder (cov) & 0.112 $\pm$ 0.004$^{***}$ & 0.609 $\pm$ 0.010$^{**}$ & 0.682 $\pm$ 0.008 & 0.400 $\pm$ 0.039$^{***}$ \\
\cline{2-5}
\textbf{$\Delta$ Best Frozen $-$ Fully-Trained} & \textbf{-7.8\%} & \textbf{-4.0\%} & \textbf{-6.0\%} & \textbf{+1.0\%} \\
\hline
\textbf{JumpReLU:} \\
\quad Fully-Trained (iso) & 0.852 $\pm$ 0.001 & 0.704 $\pm$ 0.031 & 0.843 $\pm$ 0.027 & 0.722 $\pm$ 0.046 \\
\quad Frozen Decoder (iso) & 0.457 $\pm$ 0.003$^{***}$ & 0.702 $\pm$ 0.009 & 0.771 $\pm$ 0.012 & 0.591 $\pm$ 0.027$^{*}$ \\
\quad Frozen Decoder (cov) & 0.578 $\pm$ 0.003$^{***}$ & 0.694 $\pm$ 0.015 & 0.790 $\pm$ 0.011 & 0.551 $\pm$ 0.030$^{**}$ \\
\quad Soft-Frozen Decoder (iso) & 0.789 $\pm$ 0.000$^{***}$ & 0.672 $\pm$ 0.036 & 0.735 $\pm$ 0.036 & 0.739 $\pm$ 0.038 \\
\quad Frozen Encoder (iso) & 0.609 $\pm$ 0.003$^{***}$ & 0.635 $\pm$ 0.010$^{*}$ & 0.747 $\pm$ 0.012 & 0.623 $\pm$ 0.024$^{*}$ \\
\quad Frozen Encoder (cov) & -0.017 $\pm$ 0.076$^{**}$ & 0.652 $\pm$ 0.015$^{*}$ & 0.720 $\pm$ 0.017 & 0.423 $\pm$ 0.043$^{***}$ \\
\cline{2-5}
\textbf{$\Delta$ Best Frozen $-$ Fully-Trained} & \textbf{-7.3\%} & \textbf{-0.2\%} & \textbf{-6.2\%} & \textbf{+2.3\%} \\
\hline
\end{tabular}}
\label{tab:layer12}
\end{table}

\vspace{0.001cm}

\begin{table}[h]
\caption{Frozen SAEs performance on Gemma-2-2B layer 19 L0=160.}
\centering
\resizebox{\textwidth}{!}{%
\begin{tabular}{lcccc}
\hline
\textbf{Variant} & \textbf{Explained Variance} & \textbf{Sparse Probing (top-1)} & \textbf{Sparse Probing (top-5)} & \textbf{Causal Editing} \\
\hline
\textbf{BatchTopK:} \\
\quad Fully-Trained (iso) & 0.887 $\pm$ 0.000 & 0.806 $\pm$ 0.032 & 0.900 $\pm$ 0.016 & 0.512 $\pm$ 0.011 \\
\quad Frozen Decoder (iso) & 0.404 $\pm$ 0.001$^{***}$ & 0.709 $\pm$ 0.019$^{**}$ & 0.829 $\pm$ 0.020 & 0.492 $\pm$ 0.002$^{*}$ \\
\quad Frozen Decoder (cov) & 0.555 $\pm$ 0.001$^{***}$ & 0.730 $\pm$ 0.020$^{*}$ & 0.818 $\pm$ 0.020 & 0.493 $\pm$ 0.002$^{*}$ \\
\quad Soft-Frozen Decoder (iso) & 0.812 $\pm$ 0.001$^{***}$ & 0.758 $\pm$ 0.037 & 0.823 $\pm$ 0.031 & 0.509 $\pm$ 0.010 \\
\quad Frozen Encoder (iso) & 0.562 $\pm$ 0.000$^{***}$ & 0.722 $\pm$ 0.026$^{*}$ & 0.812 $\pm$ 0.025 & 0.492 $\pm$ 0.002$^{*}$ \\
\quad Frozen Encoder (cov) & 0.523 $\pm$ 0.001$^{***}$ & 0.638 $\pm$ 0.014$^{**}$ & 0.791 $\pm$ 0.020 & 0.491 $\pm$ 0.002$^{*}$ \\
\cline{2-5}
\textbf{$\Delta$ Best Frozen $-$ Fully-Trained} & \textbf{-8.4\%} & \textbf{-6.0\%} & \textbf{-7.9\%} & \textbf{-0.5\%} \\
\hline
\textbf{JumpReLU:} \\
\quad Fully-Trained (iso) & 0.887 $\pm$ 0.000 & 0.813 $\pm$ 0.028 & 0.900 $\pm$ 0.016 & 0.511 $\pm$ 0.012 \\
\quad Frozen Decoder (iso) & 0.432 $\pm$ 0.001$^{***}$ & 0.727 $\pm$ 0.017$^{**}$ & 0.824 $\pm$ 0.022 & 0.493 $\pm$ 0.002 \\
\quad Frozen Decoder (cov) & 0.543 $\pm$ 0.000$^{***}$ & 0.727 $\pm$ 0.019$^{**}$ & 0.825 $\pm$ 0.019 & 0.493 $\pm$ 0.002 \\
\quad Soft-Frozen Decoder (iso) & 0.816 $\pm$ 0.000$^{***}$ & 0.736 $\pm$ 0.042$^{*}$ & 0.858 $\pm$ 0.026 & 0.509 $\pm$ 0.009 \\
\quad Frozen Encoder (iso) & 0.590 $\pm$ 0.001$^{***}$ & 0.701 $\pm$ 0.016$^{**}$ & 0.795 $\pm$ 0.020 & 0.494 $\pm$ 0.002 \\
\quad Frozen Encoder (cov) & 0.512 $\pm$ 0.001$^{***}$ & 0.650 $\pm$ 0.014$^{**}$ & 0.737 $\pm$ 0.019 & 0.492 $\pm$ 0.002 \\
\cline{2-5}
\textbf{$\Delta$ Best Frozen $-$ Fully-Trained} & \textbf{-7.9\%} & \textbf{-9.5\%} & \textbf{-4.7\%} & \textbf{-0.4\%} \\
\hline
\end{tabular}}
\label{tab:layer19}
\end{table}


\begin{table}[h]
\caption{Frozen SAEs performance on Llama-3-8B layer 16 L0=160.}
\centering
\resizebox{\textwidth}{!}{%
\begin{tabular}{lcccc}
\hline
\textbf{Variant} & \textbf{Explained Variance} & \textbf{Sparse Probing (top-1)} & \textbf{Sparse Probing (top-5)} & \textbf{Causal Editing} \\
\hline
\textbf{BatchTopK:} \\
\quad Fully-Trained (iso) & 0.805 $\pm$ 0.000 & 0.766 $\pm$ 0.039 & 0.874 $\pm$ 0.022 & 0.533 $\pm$ 0.015 \\
\quad Frozen Decoder (iso) & 0.268 $\pm$ 0.001$^{***}$ & 0.709 $\pm$ 0.012 & 0.779 $\pm$ 0.019 & 0.486 $\pm$ 0.004$^{*}$ \\
\quad Frozen Decoder (cov) & 0.426 $\pm$ 0.001$^{***}$ & 0.689 $\pm$ 0.014$^{*}$ & 0.771 $\pm$ 0.016 & 0.491 $\pm$ 0.005$^{*}$ \\
\quad Soft-Frozen Decoder (iso) & 0.699 $\pm$ 0.000$^{***}$ & 0.752 $\pm$ 0.046 & 0.848 $\pm$ 0.030 & 0.519 $\pm$ 0.009 \\
\quad Frozen Encoder (iso) & 0.424 $\pm$ 0.001$^{***}$ & 0.713 $\pm$ 0.010 & 0.798 $\pm$ 0.012 & 0.492 $\pm$ 0.005$^{*}$ \\
\quad Frozen Encoder (cov) & 0.408 $\pm$ 0.000$^{***}$ & 0.681 $\pm$ 0.013$^{*}$ & 0.763 $\pm$ 0.017 & 0.482 $\pm$ 0.005$^{**}$ \\
\cline{2-5}
\textbf{$\Delta$ Best Frozen $-$ Fully-Trained} & \textbf{-13.1\%} & \textbf{-1.9\%} & \textbf{-3.0\%} & \textbf{-2.7\%} \\
\hline
\textbf{JumpReLU:} \\
\quad Fully-Trained (iso) & 0.805 $\pm$ 0.000 & 0.798 $\pm$ 0.034 & 0.882 $\pm$ 0.024 & 0.527 $\pm$ 0.012 \\
\quad Frozen Decoder (iso) & 0.291 $\pm$ 0.000$^{***}$ & 0.685 $\pm$ 0.011$^{*}$ & 0.771 $\pm$ 0.015 & 0.485 $\pm$ 0.003$^{*}$ \\
\quad Frozen Decoder (cov) & 0.000 $\pm$ 0.000$^{***}$ & 0.500 $\pm$ 0.000$^{**}$ & 0.500 $\pm$ 0.000 & 0.485 $\pm$ 0.003$^{*}$ \\
\quad Soft-Frozen Decoder (iso) & 0.707 $\pm$ 0.001$^{***}$ & 0.768 $\pm$ 0.040 & 0.842 $\pm$ 0.034 & 0.521 $\pm$ 0.012 \\
\quad Frozen Encoder (iso) & 0.473 $\pm$ 0.001$^{***}$ & 0.682 $\pm$ 0.014$^{**}$ & 0.766 $\pm$ 0.011 & 0.498 $\pm$ 0.007$^{*}$ \\
\quad Frozen Encoder (cov) & 0.414 $\pm$ 0.001$^{***}$ & 0.620 $\pm$ 0.015$^{**}$ & 0.751 $\pm$ 0.009 & 0.489 $\pm$ 0.006$^{**}$ \\
\cline{2-5}
\textbf{$\Delta$ Best Frozen $-$ Fully-Trained} & \textbf{-12.1\%} & \textbf{-3.9\%} & \textbf{-4.5\%} & \textbf{-1.1\%} \\
\hline
\end{tabular}}
\label{tab:layer16}
\end{table}

\clearpage
\begingroup
\makeatletter
\setlength{\@fptop}{0pt}%
\setlength{\@fpsep}{12pt}%
\setlength{\@fpbot}{0pt plus 1fil}%
\makeatother

\begin{figure}[t]
\centering
\includegraphics[width=0.7\columnwidth]{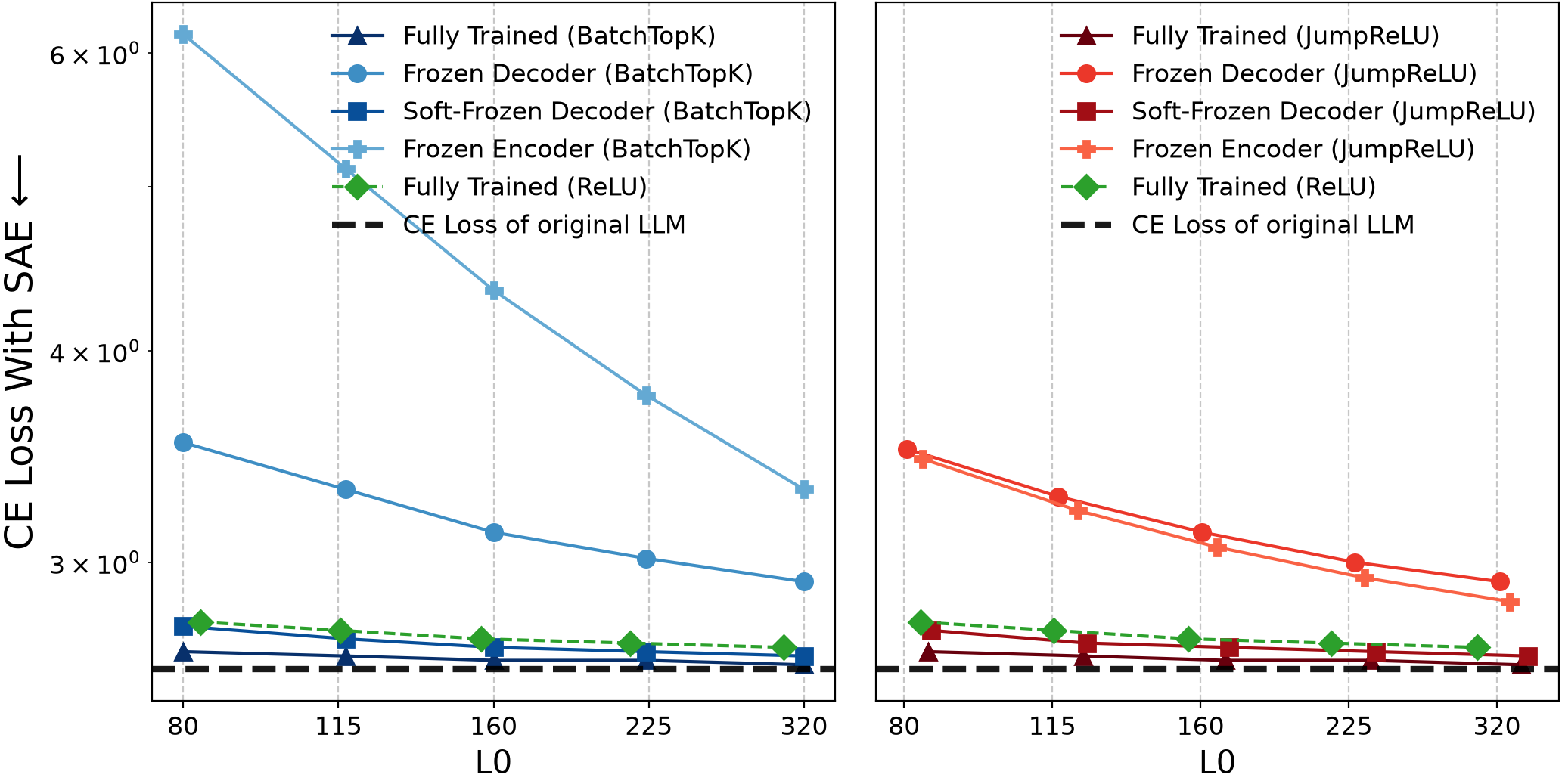}
\caption{\textbf{Cross-entropy loss comparison.} Frozen baselines match fully-trained SAEs, showing reconstruction metrics alone don't guarantee feature learning.}
\label{fig:ce}
\end{figure}

\begin{figure}[t]
\centering
\includegraphics[width=0.7\columnwidth]{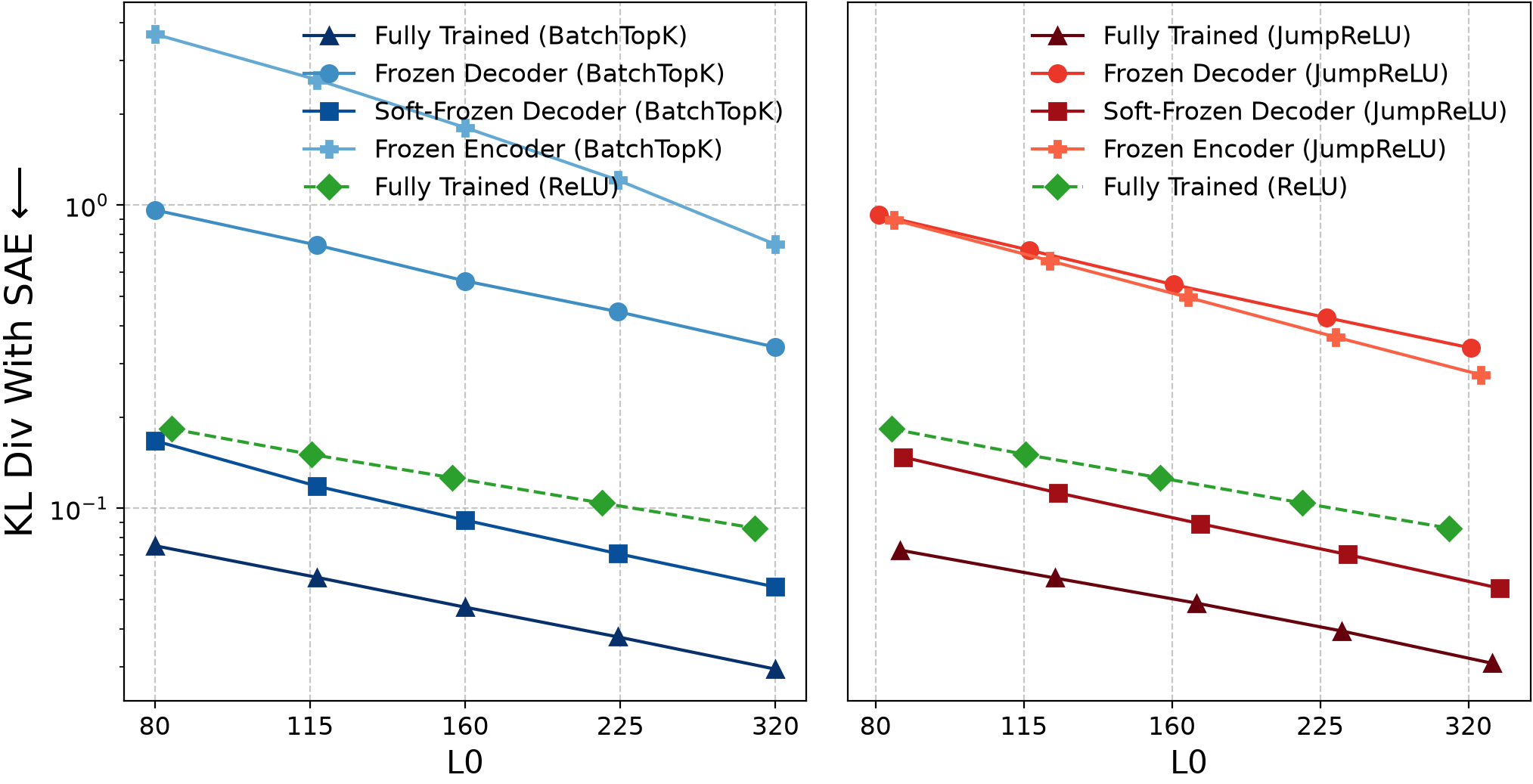}
\caption{\textbf{KL-divergence comparison.} Frozen and fully-trained SAEs produce similar KL-divergence, indicating comparable model behavior without meaningful feature learning.}
\label{fig:kl}
\end{figure}

\begin{figure}[t]
\centering
\includegraphics[width=0.7\columnwidth]{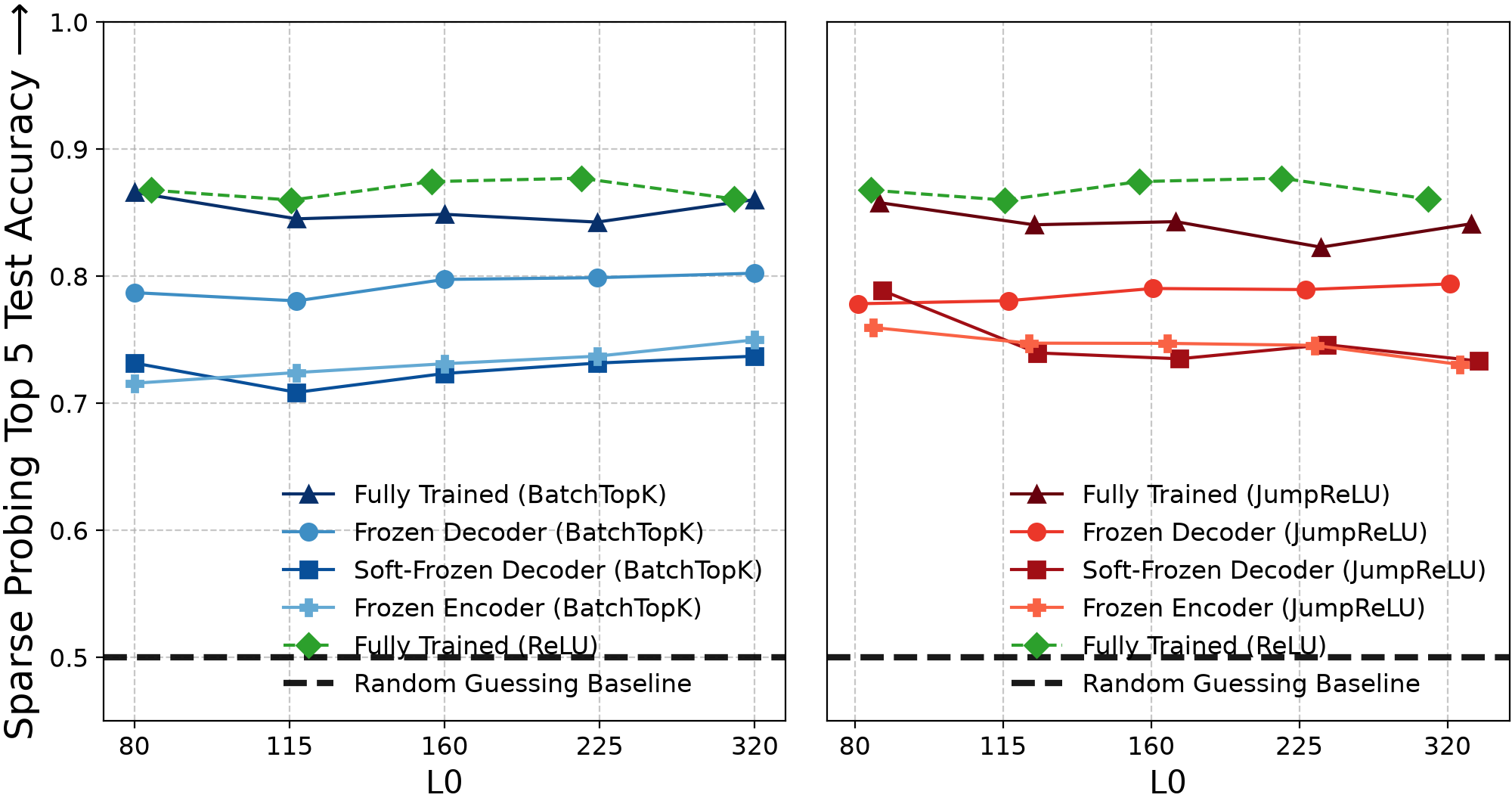}
\caption{\textbf{Sparse probing accuracy (top-5).} Frozen baselines show larger gaps vs. fully-trained SAEs, suggesting learned representations better aggregate multiple features.}
\label{fig:top5}
\end{figure}

\clearpage

  \begin{figure}[t]
    \centering
    \begin{subfigure}[b]{0.45\textwidth}
      \centering
      \includegraphics[width=\textwidth]{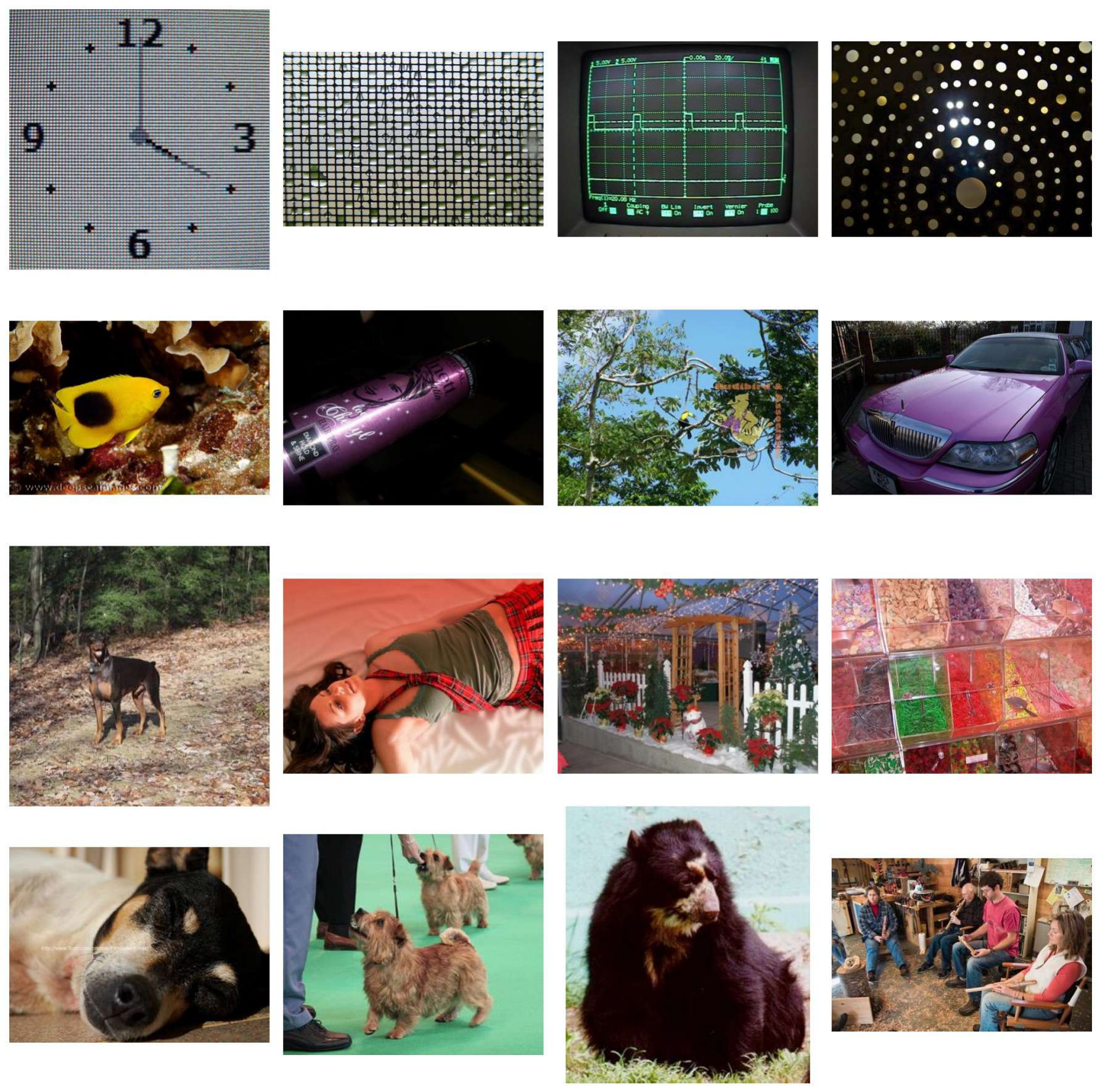}
      \caption{Random}
    \end{subfigure}
    \hfill
    \begin{subfigure}[b]{0.45\textwidth}
      \centering
      \includegraphics[width=\textwidth]{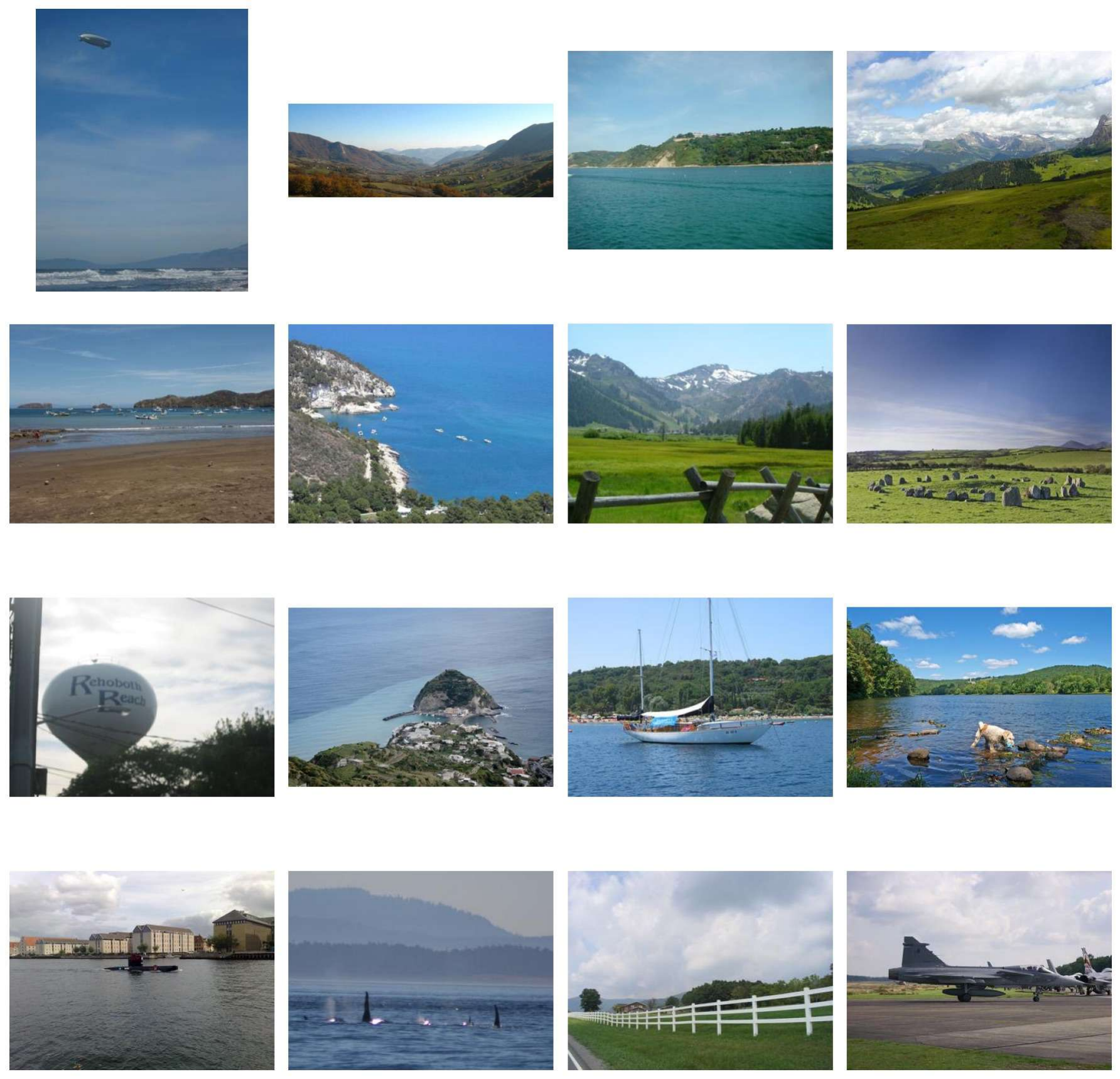}
      \caption{Trained}
    \end{subfigure}
    \caption{\textbf{Comparison of Random vs. Trained SAE Features on CLIP ViT-B/32 (Layer 3).} }
    \label{fig:clip_layer3_comparison}
  \end{figure}

  \begin{figure}[t]
    \centering
    \begin{subfigure}[b]{0.45\textwidth}
      \centering
      \includegraphics[width=\textwidth]{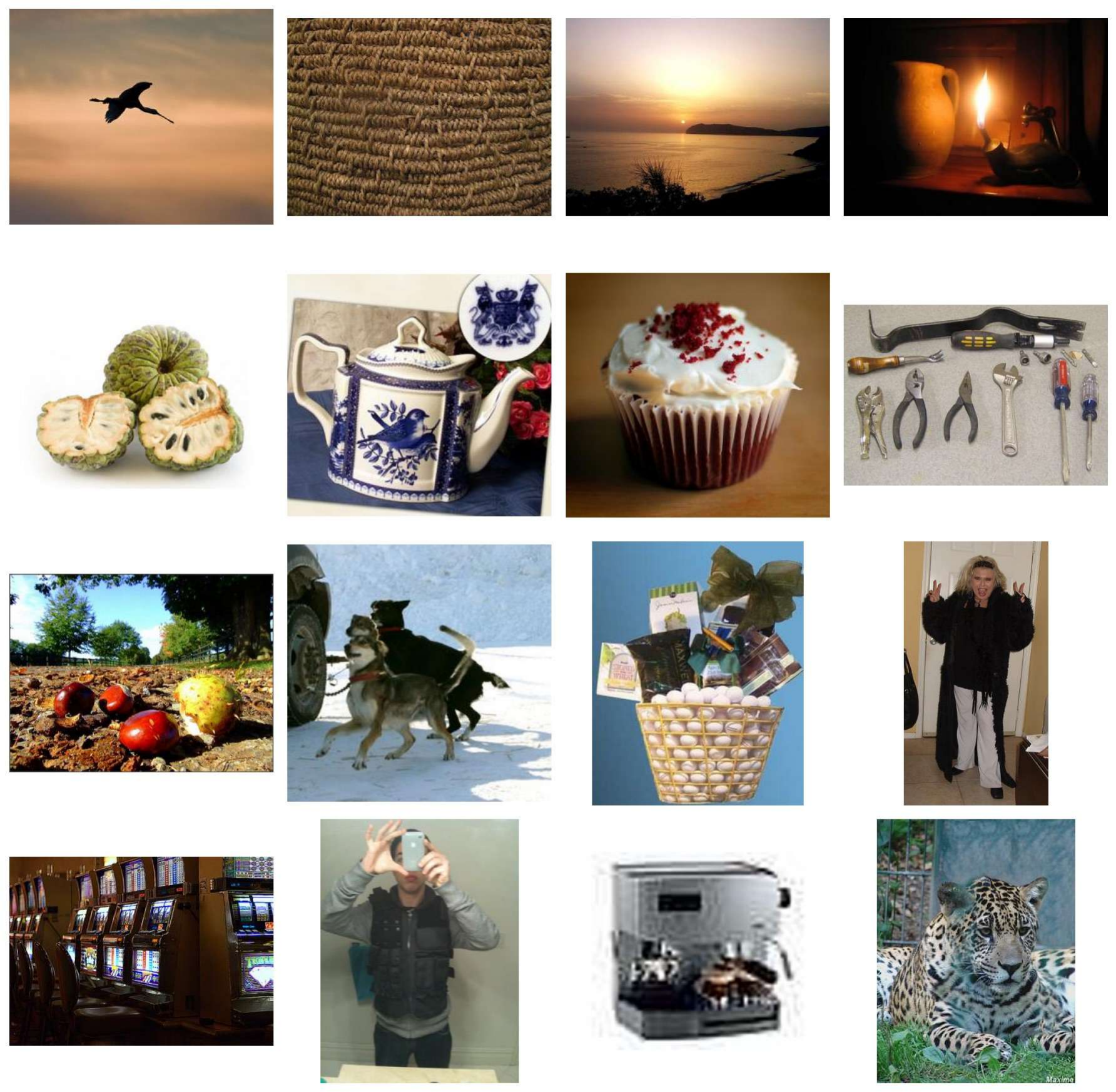}
      \caption{Random}
    \end{subfigure}
    \hfill
    \begin{subfigure}[b]{0.45\textwidth}
      \centering
      \includegraphics[width=\textwidth]{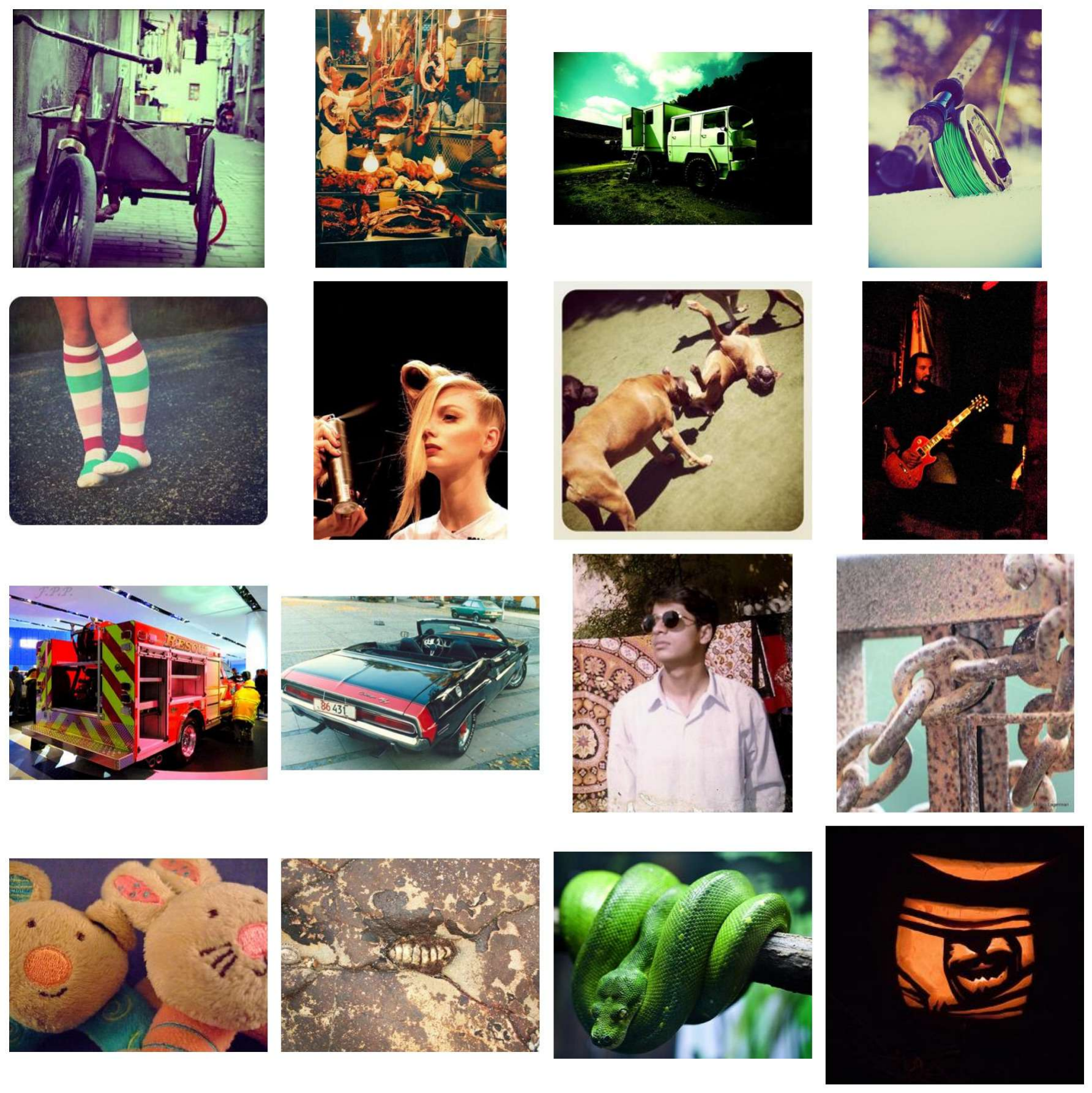}
      \caption{Trained}
    \end{subfigure}
    \caption{\textbf{Comparison of Random vs. Trained SAE Features on CLIP ViT-B/32 (Layer 5).} }
    \label{fig:clip_layer5_comparison}
  \end{figure}

\clearpage

  \begin{figure}[t]
    \centering
    \begin{subfigure}[b]{0.45\textwidth}
      \centering
      \includegraphics[width=\textwidth]{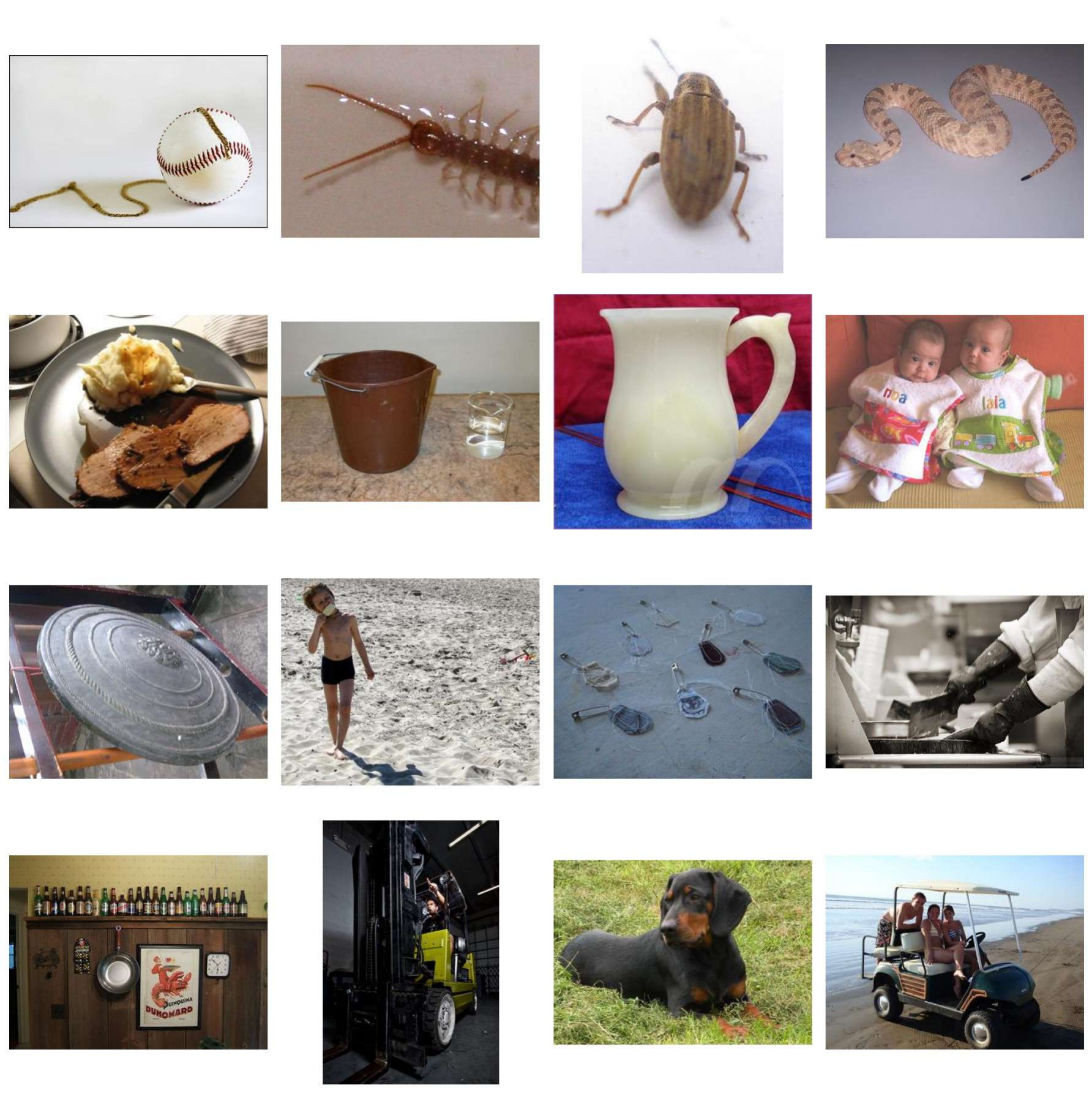}
      \caption{Random}
    \end{subfigure}
    \hfill
    \begin{subfigure}[b]{0.45\textwidth}
      \centering
      \includegraphics[width=\textwidth]{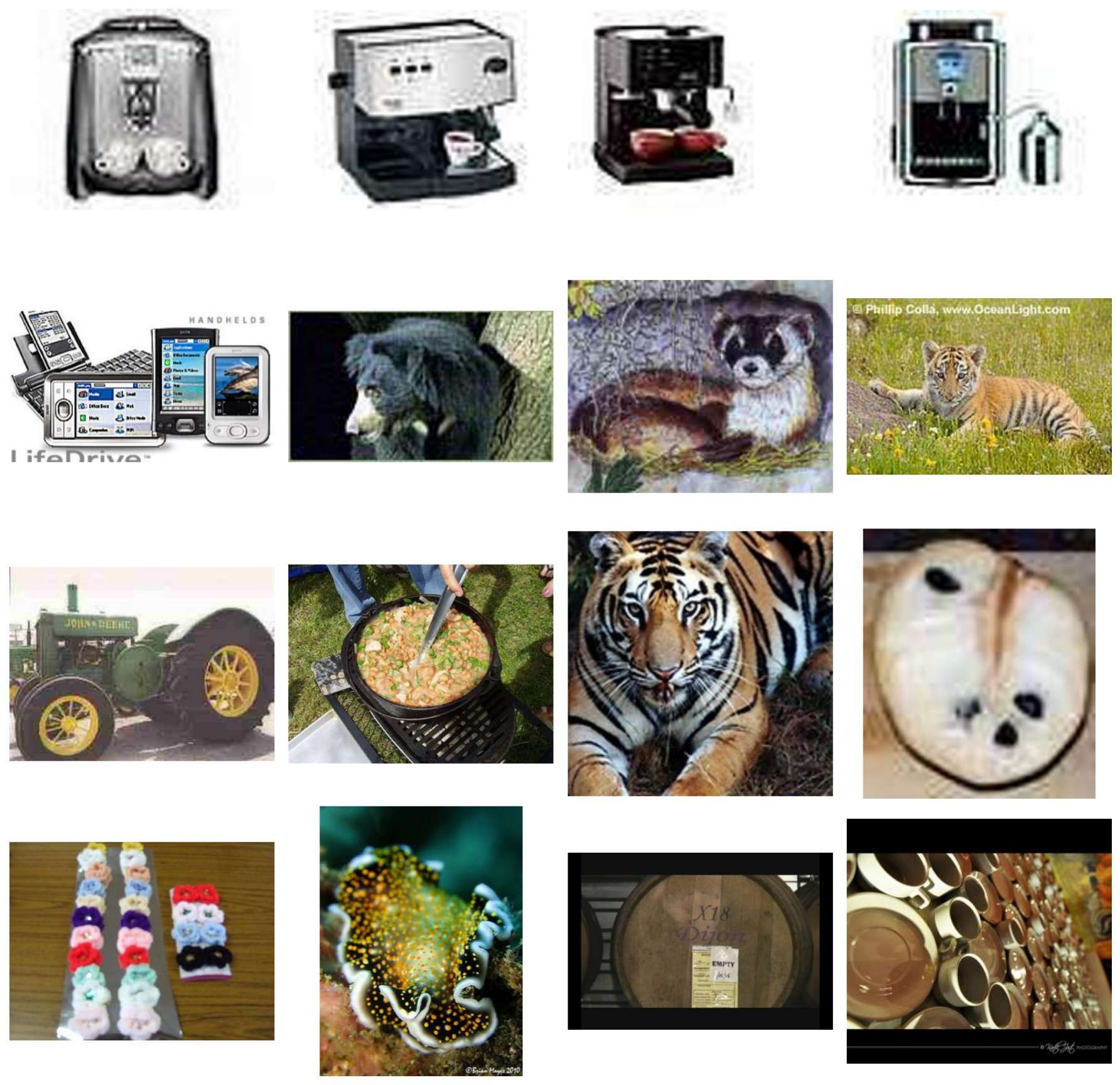}
      \caption{Trained}
    \end{subfigure}
    \caption{\textbf{Comparison of Random vs. Trained SAE Features on CLIP ViT-B/32 (Layer 7).} }
    \label{fig:clip_layer7_comparison}
  \end{figure}

  \begin{figure}[t]
    \centering
    \begin{subfigure}[b]{0.45\textwidth}
      \centering
      \includegraphics[width=\textwidth]{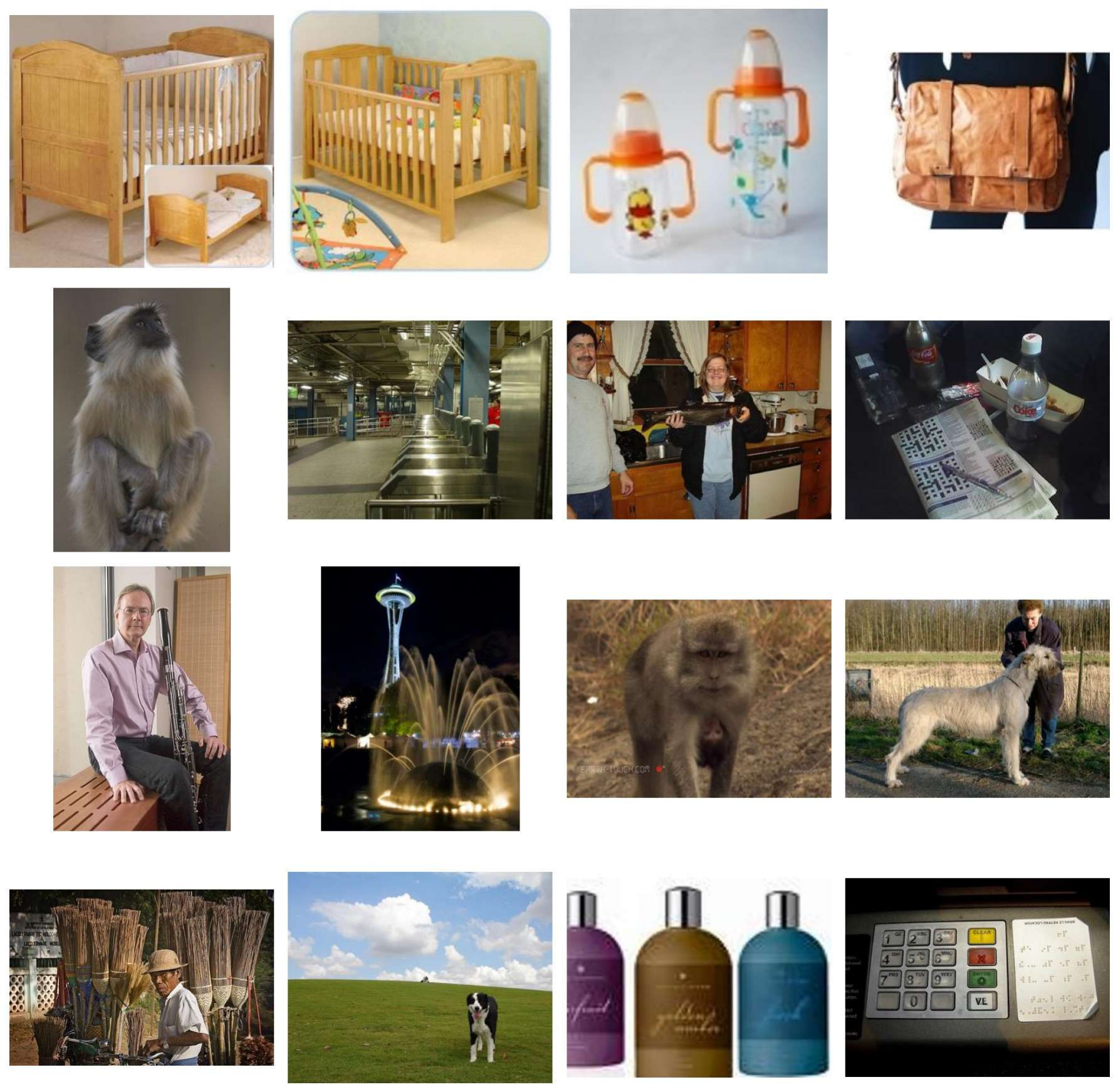}
      \caption{Random}
    \end{subfigure}
    \hfill
    \begin{subfigure}[b]{0.45\textwidth}
      \centering
      \includegraphics[width=\textwidth]{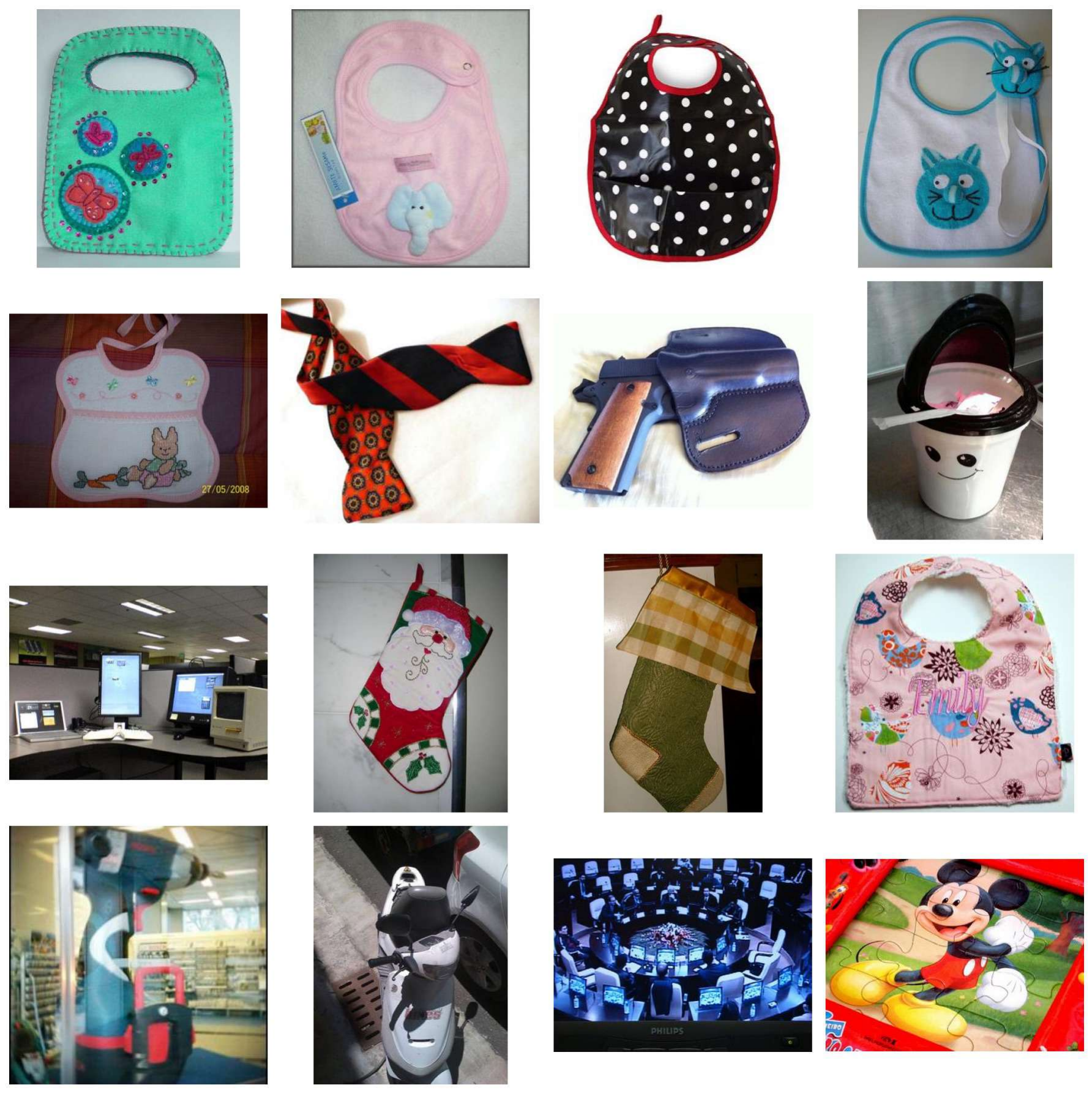}
      \caption{Trained}
    \end{subfigure}
    \caption{\textbf{Comparison of Random vs. Trained SAE Features on CLIP ViT-B/32 (Layer 9).} }
    \label{fig:clip_layer9_comparison}
  \end{figure}

\clearpage
\endgroup

\section{SynthSAEBench Results}
\label{sec:synthsaebench}

Recently, SynthSAEBench~\citep{chanin2026synthsaebench} provides a synthetic evaluation with more structured feature correlations.
We evaluated the same four SAE architectures and hyperparameters as in Section~\ref{toy_model_res} on SynthSAEBench; results confirm our findings: TopK recovers features best, while other architectures show low MMCS (mean maximum cosine similarity between true and learned features) despite high explained variance (Table~\ref{tab:synthsaebench}).

\begin{table}[h]
\caption{SynthSAEBench performance for the four SAE architectures used in our synthetic experiments. MMCS measures feature recovery.}
\label{tab:synthsaebench}
\centering
\small
\begin{tabular}{lccc}
\hline
\textbf{Architecture} & \textbf{$k$} & \textbf{EV} $\uparrow$ & \textbf{MMCS} $\uparrow$ \\
\hline
TopK & 20 & 0.897 & 0.644 \\
TopK & 100 & 0.947 & 0.825 \\
BatchTopK & 20 & 0.799 & 0.322 \\
BatchTopK & 100 & 0.891 & 0.489 \\
JumpReLU & 20 & 0.844 & 0.441 \\
JumpReLU & 100 & 0.889 & 0.499 \\
Matryoshka & 20 & 0.773 & 0.309 \\
Matryoshka & 100 & 0.840 & 0.375 \\
\hline
\end{tabular}
\end{table}

\section{A Minimal Evaluation Protocol for New SAE Architectures}
\label{sec:protocol}

We recommend the following low-barrier checklist when claiming that standard metrics certify meaningful feature learning. Reference implementations of all three baselines will be included in our code release.

\begin{enumerate}
    \item Train three frozen baselines alongside the candidate under identical data and hyperparameters, using \textit{cov} initialization for Frozen Decoder and \textit{iso} for Frozen Encoder and Soft-Frozen Decoder.
    \item Evaluate the candidate and all three baselines on the same metrics used to argue meaningful feature learning.
    \item Use multiple seeds where feasible to confirm that gaps vs.\ baselines are stable.
    \item Read gaps per metric rather than as a single pass/fail verdict.
\end{enumerate}

\section*{NeurIPS Paper Checklist}

\begin{enumerate}

\item {\bf Claims}
    \item[] Question: Do the main claims made in the abstract and introduction accurately reflect the paper's contributions and scope?
    \item[] Answer: \answerYes{}
    \item[] Justification: All claims in the abstract and introduction (synthetic failure, three baselines, baselines matching SAEs) are supported by results in Sections~3 and~4.

\item {\bf Limitations}
    \item[] Question: Does the paper discuss the limitations of the work performed by the authors?
    \item[] Answer: \answerYes{}
    \item[] Justification: Section~5 (Limitations) discusses that the synthetic setup assumes independent feature activations, omitting correlations present in real networks.

\item {\bf Theory assumptions and proofs}
    \item[] Question: For each theoretical result, does the paper provide the full set of assumptions and a complete (and correct) proof?
    \item[] Answer: \answerNA{}
    \item[] Justification: The paper does not include formal theoretical results; Appendix~D contains a supporting probabilistic argument but not a theorem in the traditional sense.

    \item {\bf Experimental result reproducibility}
    \item[] Question: Does the paper fully disclose all the information needed to reproduce the main experimental results of the paper to the extent that it affects the main claims and/or conclusions of the paper (regardless of whether the code and data are provided or not)?
    \item[] Answer: \answerYes{}
    \item[] Justification: Section~4.1 provides all training details: model, layer, optimizer, learning rate, batch size, dataset, context length, expansion factor, and sparsity levels. Code is available at \url{https://github.com/AntonKorznikov/sanity-checks-saes} and trained SAEs will be publicly released.

\item {\bf Open access to data and code}
    \item[] Question: Does the paper provide open access to the data and code, with sufficient instructions to faithfully reproduce the main experimental results, as described in supplemental material?
    \item[] Answer: \answerYes{}
    \item[] Justification: Code is available at the public repository \url{https://github.com/AntonKorznikov/sanity-checks-saes}. Trained SAEs will be released upon acceptance.

\item {\bf Experimental setting/details}
    \item[] Question: Does the paper specify all the training and test details (e.g., data splits, hyperparameters, how they were chosen, type of optimizer) necessary to understand the results?
    \item[] Answer: \answerYes{}
    \item[] Justification: Section~4.1 specifies optimizer, learning rate, batch size, dataset, context length, expansion factor, sparsity levels, and model layers for all experiments.

\item {\bf Experiment statistical significance}
    \item[] Question: Does the paper report error bars suitably and correctly defined or other appropriate information about the statistical significance of the experiments?
    \item[] Answer: \answerYes{}
    \item[] Justification: Tables~1--3 in the appendix report mean $\pm$ standard deviation across runs for all metrics.

\item {\bf Experiments compute resources}
    \item[] Question: For each experiment, does the paper provide sufficient information on the computer resources (type of compute workers, memory, time of execution) needed to reproduce the experiments?
    \item[] Answer: \answerYes{}
    \item[] Justification: Section~4.1 states that all training used a single NVIDIA A100-80GB GPU, with each model requiring $\approx$ 12 hours on 500M tokens;
    
\item {\bf Code of ethics}
    \item[] Question: Does the research conducted in the paper conform, in every respect, with the NeurIPS Code of Ethics \url{https://neurips.cc/public/EthicsGuidelines}?
    \item[] Answer: \answerYes{}
    \item[] Justification: The paper performs evaluation of existing methods on publicly available models and datasets; no human subjects or sensitive data are involved.

\item {\bf Broader impacts}
    \item[] Question: Does the paper discuss both potential positive societal impacts and negative societal impacts of the work performed?
    \item[] Answer: \answerNA{}
    \item[] Justification: The paper is a methodological evaluation of interpretability tools with no direct path to harmful applications.
    
\item {\bf Safeguards}
    \item[] Question: Does the paper describe safeguards that have been put in place for responsible release of data or models that have a high risk for misuse (e.g., pre-trained language models, image generators, or scraped datasets)?
    \item[] Answer: \answerNA{}
    \item[] Justification: The released assets (SAE checkpoints and evaluation code) pose no meaningful misuse risk.

\item {\bf Licenses for existing assets}
    \item[] Question: Are the creators or original owners of assets (e.g., code, data, models), used in the paper, properly credited and are the license and terms of use explicitly mentioned and properly respected?
    \item[] Answer: \answerYes{}
    \item[] Justification: All assets (OpenWebText, Gemma-2-2B, Llama-3-8B, CLIP, SAEBench, RAVEL) are cited with their original papers; all are publicly available under open research licenses.

\item {\bf New assets}
    \item[] Question: Are new assets introduced in the paper well documented and is the documentation provided alongside the assets?
    \item[] Answer: \answerYes{}
    \item[] Justification: The public repository at \url{https://github.com/AntonKorznikov/sanity-checks-saes} includes documentation. Trained SAEs will be released with full hyperparameters and instructions.

\item {\bf Crowdsourcing and research with human subjects}
    \item[] Question: For crowdsourcing experiments and research with human subjects, does the paper include the full text of instructions given to participants and screenshots, if applicable, as well as details about compensation (if any)?
    \item[] Answer: \answerNA{}
    \item[] Justification: The paper does not involve crowdsourcing or human subjects.

\item {\bf Institutional review board (IRB) approvals or equivalent for research with human subjects}
    \item[] Question: Does the paper describe potential risks incurred by study participants, whether such risks were disclosed to the subjects, and whether Institutional Review Board (IRB) approvals (or an equivalent approval/review based on the requirements of your country or institution) were obtained?
    \item[] Answer: \answerNA{}
    \item[] Justification: The paper does not involve human subjects.

\item {\bf Declaration of LLM usage}
    \item[] Question: Does the paper describe the usage of LLMs if it is an important, original, or non-standard component of the core methods in this research? Note that if the LLM is used only for writing, editing, or formatting purposes and does \emph{not} impact the core methodology, scientific rigor, or originality of the research, declaration is not required.
    \item[] Answer: \answerYes{}
    \item[] Justification: GPT-4o-mini is a core part of the AutoInterp evaluation pipeline (Section~4.1): it generates feature descriptions and scores them via binary classification.

\end{enumerate}

\end{document}